\RequirePackage{fix-cm}
\documentclass[twocolumn]{svjour3}          
\smartqed  
\usepackage{graphicx}
\usepackage[numbers]{natbib}
\usepackage{multirow}
\usepackage{amsmath}
\usepackage{amssymb}
\usepackage{bbm}
\usepackage{bm}
\usepackage{booktabs}
\usepackage{array} 
\usepackage{blindtext}
\usepackage{comment}

\usepackage{xspace}

\usepackage[utf8]{inputenc} 
\usepackage[T1]{fontenc}    
\usepackage{booktabs}       
\usepackage{amsfonts}       
\usepackage{nicefrac}       
\usepackage{microtype}      
\usepackage{xcolor}         

\usepackage{graphicx}
\usepackage{subfigure}
\usepackage{amsmath,amssymb} 
\usepackage{caption} 
\usepackage{hyperref}
\usepackage{wrapfig}
\usepackage[american]{babel}
\usepackage{subfigure}
\usepackage{csquotes}

\usepackage{bm}
\usepackage{paralist,algorithm}
\usepackage{subfigure}
\usepackage{multirow}

\usepackage[noend]{algpseudocode}

\newcommand{\eg}{{\em e.g.}}
\newcommand{\ie}{{\em i.e.}}
\newcommand{\name}{{\sc Aper }}
\newcommand{\mame}{{\sc Aper}}
\newcommand{\bfname}[1]{{\bf #1}}

\newcommand{\x}{{\bf x}}

\newcommand{\p}{{\bf p}}

\newcommand{\w}{{\bf w}}

\newcommand{\R}{\mathbb{R}}

\newcommand{\D}{\mathcal{D}}

\usepackage{accents}

\definecolor{Gray}{gray}{0.85}
\definecolor{LightCyan}{rgb}{0.88,1,1}

\usepackage{subfigure}

\makeatletter
\DeclareRobustCommand\onedot{\futurelet\@let@token\@onedot}
\def\@onedot{\ifx\@let@token.\else.\null\fi\xspace}

\def\eg{\emph{e.g}\onedot} 
\def\ie{\emph{i.e}\onedot}

\makeatother

\usepackage{pifont}

\usepackage[misc]{ifsym}
\usepackage{color, colortbl}
\definecolor{citecolor}{HTML}{0071bc}
\definecolor{tabhighlight}{HTML}{e5e5e5}

\makeatletter
\renewcommand\paragraph{
  \@startsection{paragraph} 
  {4} 
  {\z@} 
  {.5em \@plus1ex \@minus.2ex} 
  {-.5em} 
  {\normalfont\normalsize\bfseries} 
}
\makeatother

\begin{document}
\sloppy

\title{Revisiting Class-Incremental Learning with Pre-Trained Models: Generalizability and Adaptivity are All You Need 
}

\author{Da-Wei Zhou  \and
        Zi-Wen Cai \and
        Han-Jia Ye \and
        De-Chuan Zhan  \and
        Ziwei Liu
}

\institute{Da-Wei Zhou \at
            School of Artificial Intelligence, National Key Laboratory for Novel Software Technology, Nanjing University, China \\
              \email{zhoudw@lamda.nju.edu.cn}
           \and
           Zi-Wen Cai \at
             School of Artificial Intelligence, National Key Laboratory for Novel Software Technology, Nanjing University, China \\
              \email{caizw@lamda.nju.edu.cn}
           \and
           Han-Jia Ye \at
            School of Artificial Intelligence, National Key Laboratory for Novel Software Technology, Nanjing University, China \\
              \email{yehj@lamda.nju.edu.cn}
           \and
           De-Chuan Zhan \at
            School of Artificial Intelligence, National Key Laboratory for Novel Software Technology, Nanjing University, China \\
              \email{zhandc@nju.edu.cn}
           \and
           Ziwei Liu \at
           S-Lab, Nanyang Technological University, Singapore \\
           \email{ziwei.liu@ntu.edu.sg}
           \and 
           Han-Jia Ye and Ziwei Liu are corresponding authors. \\
           Work done when Da-Wei Zhou was a visiting scholar at NTU.
}

\date{Received: date / Accepted: date}

\maketitle

\begin{abstract}
   	Class-incremental learning (CIL) aims to adapt to emerging new classes without forgetting old ones. 
   Traditional CIL models are trained from scratch to continually acquire knowledge as data evolves.
   Recently, pre-training has achieved substantial progress, making vast pre-trained models (PTMs) accessible for CIL. 
   Contrary to traditional methods, PTMs possess generalizable embeddings, which can be easily transferred for CIL.
   In this work, we revisit CIL with PTMs and argue that the core factors in CIL are adaptivity for model updating and generalizability for knowledge transferring. 
   \textbf{1)} We first reveal that frozen PTM can already provide generalizable embeddings for CIL.
   Surprisingly, a simple baseline (\textbf{SimpleCIL}) which continually sets the classifiers of PTM to prototype features can beat state-of-the-art even without training on the downstream task. 
   \textbf{2)} Due to the distribution gap between pre-trained and downstream datasets, PTM can be further cultivated with adaptivity via model adaptation. We propose \textbf{AdaPt and mERge (\mame)}, which aggregates the embeddings of PTM and adapted models for classifier construction. 
   \name is a general framework that can be orthogonally combined with any parameter-efficient tuning method, which holds the advantages of PTM's generalizability and adapted model's adaptivity. 
   \textbf{3)} Additionally, considering previous ImageNet-based benchmarks are unsuitable in the era of PTM due to data overlapping, we propose four new benchmarks for assessment, namely ImageNet-A, ObjectNet, OmniBenchmark, and VTAB. Extensive experiments validate the effectiveness of \name with a unified and concise framework. 
   Code is available at \url{https://github.com/zhoudw-zdw/RevisitingCIL}.
\end{abstract}

\keywords{Class-Incremental Learning, Pre-Trained Models, Continual Learning, Catastrophic Forgetting}

\section{Introduction}

With the advancement of deep learning, deep models have achieved impressive feats in many fields~\cite{zhong2021neighborhood,liu2017cross,akbari2023alternating,wang2019learning,jaimes2007multimodal,ning2024moiretracker,yang2002detecting,ning2023rf}. However, most research focuses on recognizing a limited number of classes in static environments. In the real world, applications often deal with streaming data with incoming new classes~\cite{gomes2017survey}. To address this issue, Class-Incremental Learning (CIL) has been proposed, which allows the model to learn from the evolving data and {\em continuously} build a unified classification model. 
Nevertheless, when new classes are added sequentially, the notorious {\em catastrophic forgetting} occurs~\cite{french1999catastrophic}, which erases the previously learned knowledge. 
While typical CIL methods assume that the model is ``{\em trained from scratch},''
recent advancements in pre-training~\cite{han2021pre} have made Pre-Trained Models (PTMs) more accessible for designing models in downstream tasks. 
These PTMs are often trained on massive corpus~\cite{radford2021learning} or abundant images~\cite{deng2009imagenet,ridnik2021imagenet} with handcrafted tricks~\cite{steiner2021train}, resulting in strong {\em generalizability}. 
Consequently, several methods~\cite{wang2022learning,wang2022dualprompt,wang2022s,villa2022pivot} propose to leverage PTM for better incremental learning.

Powerful PTMs alleviate the burden of CIL~\cite{zhou2023class}. However, upon revisiting the objective of CIL, we find 
essential differences between these protocols. 
Without PTMs, CIL models are trained from {\em random initialization} to {\em continually acquire} the knowledge of new classes and build a unified embedding space, which requires the {\bf adaptivity} for sequential updating. 
In contrast, PTMs are trained with massive datasets, which makes it easier to achieve a powerful embedding space with strong {\bf generalizability}.
To use a human learning analogy,
non-PTM methods aim to teach an {\em infant} to grow up and continually acquire knowledge through college, while PTM-based methods teach an experienced {\em adult} to do the same thing, which is much easier.

\begin{figure}[t]
	\centering
	{
		\includegraphics[width=1\columnwidth]{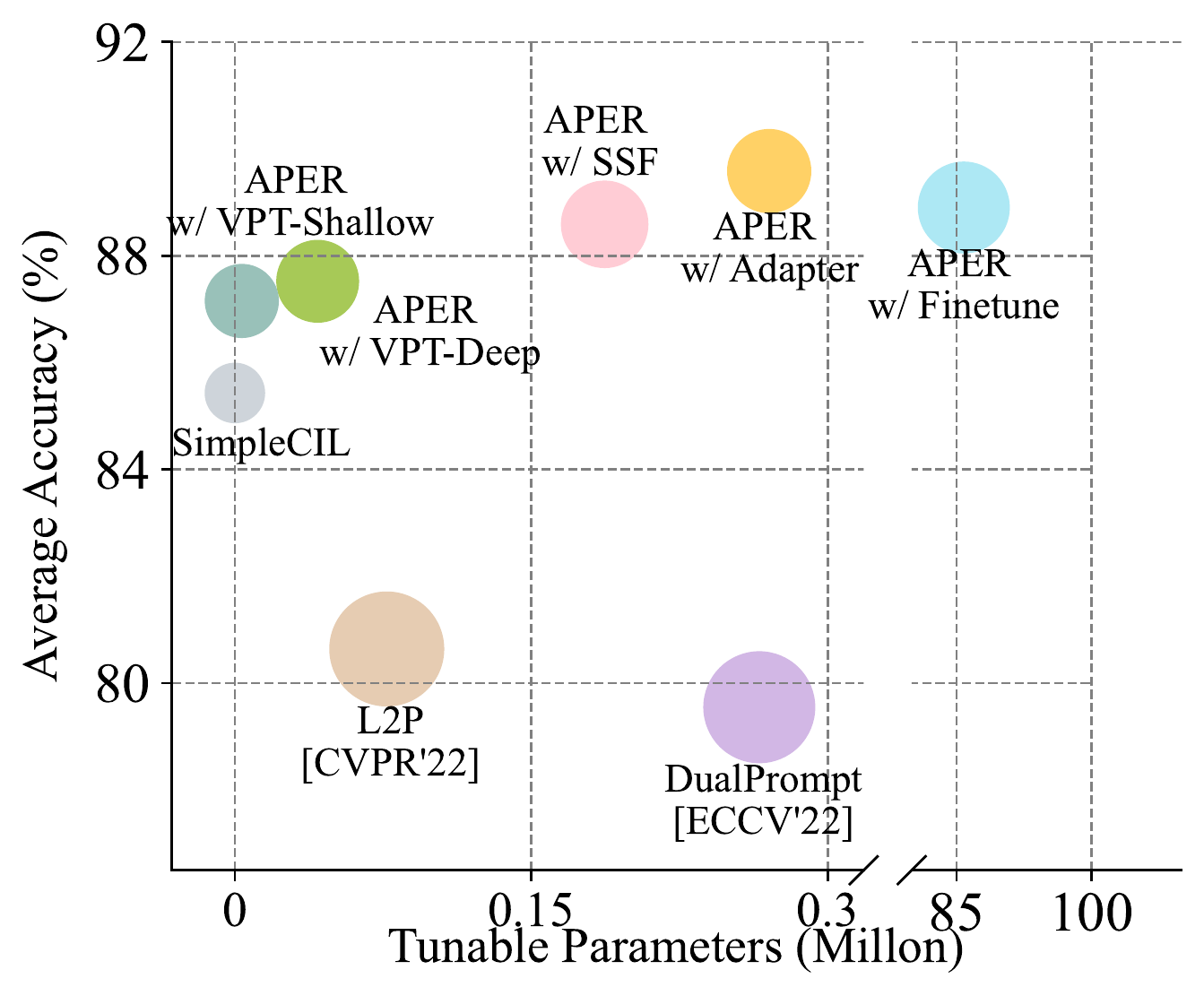}
	}
	\vspace{-4mm}
	\caption{ \small Comparison of different PTM-based CIL methods on VTAB dataset. The X-axis stands for the number of tunable parameters, and the Y-axis represents the average accuracy. The radius stands for the training time. Although consuming more tuning parameters and training time, current state-of-the-art (\ie, L2P and DualPrompt) still show inferior performance than the baseline method SimpleCIL. By contrast, our \name  consistently improves the baseline with tiny costs.
		For a fair comparison, all methods are based on pre-trained ViT-B/16-IN1K. SimpleCIL utilizes the training set to calculate the average embeddings.
		\vspace{-4mm}
	} \label{figure:intro}
\end{figure}
To evaluate the generalizability of PTMs, we formulate a CIL task using the VTAB~\cite{zhai2019large} dataset and test the performance of state-of-the-art PTM-based methods~\cite{wang2022dualprompt,wang2022learning} with a pre-trained ViT-B/16-IN1K in Figure~\ref{figure:intro}. 
As a comparison, we present a simple baseline {\bf SimpleCIL} to evaluate the quality of the pre-trained features. 
With the pre-trained embedding function frozen, SimpleCIL sets the classifier weights to the average embeddings~\cite{snell2017prototypical} of each new class for classification. 
The average embeddings are calculated with training images.
In this way, SimpleCIL serves as a direct indicator of the quality of pre-trained features. If PTMs possess generalizable features, directly matching the average pattern to each query instance could also achieve competitive results.
Surprisingly, Figure 1 shows that SimpleCIL outperforms the current SOTA by 5\% even without any tuning on these downstream tasks, verifying its strong {\em generalizability} in knowledge transfer.

Although PTMs are generalizable for CIL, a {\em domain gap} may still exist between pre-trained and incremental datasets~\cite{zhou2022domain,you2020co}.
For instance, the ImageNet pre-trained model may not generalize well to out-of-distribution~\cite{hendrycks2021natural} or specialized tasks~\cite{alfassy2022feta}. Under such circumstances, freezing the embedding for knowledge transferring is not a ``{\em panacea}.'' Accordingly, {\em adaptivity} becomes essential to enable the model to grasp {\em task-specific features}. Nevertheless, sequentially tuning the PTM will harm the structural information and weaken the generalizability~\cite{kumarfine}, leading to the irreversible forgetting of previous knowledge. Is there a way to {\em unify} the generalizability of PTM with the adaptivity of the adapted model?

In this paper, we present {\bf AdaPt and mERge} (\mame) for CIL, which employs PTM to enhance generalizability and adaptivity in a unified framework. To improve adaptivity, we adapt the PTM in the first incremental stage via parameter-efficient tuning. Adapting the model helps to obtain task-specific features and fills the domain gap between PTM and incremental data.
We then concatenate the adapted model with the PTM to extract average embeddings as the classifier, thereby maintaining generalizability.
\name restricts model tuning in the first stage, striking a balance between adaptivity and generalizability. 
Moreover, typical ImageNet-based CIL benchmarks are unsuitable for evaluation due to overlapping between pre-trained and downstream tasks.
Therefore, we benchmark PTM-based CIL with four new datasets that have {\em large} domain gaps with the pre-trained data. Extensive experiments under various settings demonstrate the effectiveness of \mame.
Our main contributions can be summarized as follows:

\begin{itemize}
	\item With extensive empirical evaluations, we reveal a simple baseline (\ie, SimpleCIL) can directly transfer the strong generalizability of PTMs in CIL, which even outperforms current state-of-the-art without training on the downstream task.
	\item We propose \name that employs PTM to enhance generalizability and adaptivity in a unified framework. It enjoys the generalizability of PTMs using a prototype-based classifier and the strong adaptivity of downstream tasks by model adaptation. Due to its uniformity, \name can be applied to different network structures and different tuning techniques;
	\item  Due to the overlapping between pre-trained data and traditional CIL benchmarks, we benchmark pre-trained model-based CIL with several new datasets with large domain gaps with ImageNet. Extensive experiments on these benchmark datasets verify \mame's state-of-the-art performance.
\end{itemize}

\section{Related Work}

\subsection{Class-Incremental Learning (CIL)}
Class-incremental learning enables a learning system to continually incorporate new concepts without forgetting old ones~\cite{zhou2023class,wang2023comprehensive,masana2022class,zhang2023vision,li2023dpps,cai2024single,yang2024exploring,li2023configure,ding2023structural,zheng2023preserving,zheng2024multi}. Typical CIL methods can be divided into several categories. Exemplar-based methods save and replay exemplars from old classes to recover former knowledge~\cite{aljundi2019gradient,chaudhry2018riemannian,iscen2020memory,liu2020mnemonics}.
Apart from direct saving exemplars, other methods work on saving features~\cite{zhao2021memory,iscen2020memory,zhu2021prototype} or using generative models~\cite{shin2017continual,jiang2021ib,smith2021always,gao2023ddgr} to construct the memory.
Knowledge distillation-based methods aim to align the outputs of old and new models during updating, thereby maintaining knowledge of old concepts~\cite{li2017learning,rebuffi2017icarl,douillard2020podnet,zhang2020class,hu2021distilling}. The alignment can be built in several aspects, resulting in different optimization targets. iCaRL~\cite{rebuffi2017icarl} and LwF~\cite{li2017learning} utilize logit distillation, which requires the output logits on old classes to be the same. LUCIR~\cite{hou2019learning} utilizes feature distillation and forces the output features to be the same across models. Some following works distill other feature products to resist forgetting, \eg, attention map~\cite{dhar2019learning}, weighted feature map~\cite{kang2022class}, pooled features~\cite{douillard2020podnet}, casual effect~\cite{hu2021distilling}, subspace feature~\cite{simon2021learning}, and spatial/temporal features~\cite{zhao2021video}.
Additionally, other works also consider distilling the relational information among a group of instances~\cite{dong2021few,gao2022rdfcil,tao2020few,dong2023heterogeneous}.
The third group finds the inductive bias in the incremental model and designs rectification algorithms for an unbiased prediction. BiC~\cite{wu2019large} and IL2M~\cite{belouadah2019il2m} calibrate the logit scales between old and new classes to resist forgetting former classes. WA~\cite{zhao2020maintaining} directly normalizes the fully connected layer to alleviate its influence on the final prediction. SDC~\cite{yu2020semantic} estimates the prototype drift of former classes via new class instances.
TEEN~\cite{wang2024few} designs a prototype calibration process to adjust the few-shot prototypes of new classes for better classification.
The following works also consider rectifying the biased BN statistics~\cite{pham2021continual} and feature representations~\cite{shi2022mimicking}. Recently, network expansion-based methods have shown competitive performance, which can be further divided into neuron-wise, backbone-wise, and token-wise. Neuron-wise~\cite{yoon2018lifelong,xu2018reinforced} expansion aims tom to expand the network's width to enhance its representation ability. Backbone-wise~\cite{yan2021dynamically,wang2022foster,zhou2022model,wang2023beef} expansion methods aim to build a holistic embedding by training a separate backbone for each new task and aggregate them as the final representation. Finally, token-wise~\cite{douillard2022dytox,wang2022dualprompt,wang2022learning,wang2022s} expansion are designed to add lightweight tokens to adapt the model while preserving its knowledge. CIL algorithms are also widely adopted in other real-world applications, \eg, federated learning~\cite{dong2022federated}, semantic segmentation~\cite{cermelli2022incremental}, text-to-image diffusion~\cite{sun2024create}, and object detection~\cite{dong2021i3dol,perez2020incremental}.

\subsection{CIL with Pre-Trained Models}
Pre-trained model-based CIL~\cite{zhou2024continual} is becoming popular with the increasing prevalence of pre-trained models~\cite{dosovitskiy2020image,radford2021learning}. The aim is to sequentially adjust the PTM to stream data with new classes without forgetting. 
L2P~\cite{wang2022learning} applies visual prompt tuning~\cite{jia2022visual} to CIL based on the pre-trained Vision Transformer~\cite{dosovitskiy2020image} and learns a prompt pool to select the instance-specific prompt. 
During training, L2P retrieves the nearest prompts to the query instance and appends them to get the instance-specific embedding.
DualPrompt~\cite{wang2022dualprompt} extends L2P with general and expert prompts. Specifically, general prompts are equally assigned to all tasks, while expert prompts are selected for the specific task via the prompt retrieval process.  
Unlike the key-value search in L2P, CODA-Prompt~\cite{smith2023coda} improves the prompt selection process with an attention mechanism so that the reweighted prompt can reflect the task-specific information of all seen classes.
Furthermore, DAP~\cite{jung2023generating} learns a prompt generator that can generate instance-specific prompts instead of the complex prompt retrieval process.
SLCA~\cite{zhang2023slca} explores the classifier rectification process~\cite{zhu2021prototype} during model updating.
ESN~\cite{wang2022isolation} adopts the anchor-based energy self-normalization strategy to aggregate multiple pre-trained classifiers. 
CPP~\cite{li2024steering} designs task-specific prompt-tuning with a contrastive learning objective.
LAE~\cite{gao2023unified} utilizes exponential moving average (EMA) among online and offline models to resist forgetting. Although it also considers learning multiple models for CIL, our work differs from it in the inference format and the updating policy. 
Apart from the single modality for visual recognition, recent research also involves incremental learning of pre-trained vision-language models~\cite{yu2022coca,yuan2021florence,tschannen2022image}.
When changing ViT into CLIP~\cite{radford2021learning}, S-Prompts~\cite{wang2022s} and Pivot~\cite{villa2022pivot} extend L2P by learning prompts for both text and image modalities~\cite{zhou2022learning}.
A contemporary work~\cite{liu2023large} explores the application of PTMs in class-incremental novel class discovery~\cite{roy2022class}, which shows a frozen PTM can detect and learn new classes with high performance. However, apart from the frozen PTM, this manuscript also explores the effect of downstream data in adapting the model, aiming to unify the generalizability of PTM and adaptivity of the downstream data.

\subsection{Parameter-Efficient Tuning for Pre-Trained Models}
Parameter-efficient tuning aims to adapt the pre-trained model to downstream tasks by tuning only a small number of (extra) parameters. Compared to fully finetuning, parameter-efficient tuning obtains competitive or even better performance at a much lower cost. Visual prompt tuning (VPT)~\cite{jia2022visual} prepends tunable prefix tokens~\cite{li2021prefix} to the input or hidden layers. LoRA~\cite{hulora} learns low-rank matrices to approximate parameter updates. AdaptFormer~\cite{chenadaptformer} learns extra adapter~\cite{rebuffi2017learning} modules with downsize and upsize projection. 
AdapterFusion~\cite{pfeiffer2021adapterfusion} merges the learned adapters with a fusion module. SSF~\cite{lianscaling} addresses the scaling and shifting operation for model tuning. 
BitFit~\cite{zaken2021bitfit} only tunes the bias term in the pre-trained model, and FacT~\cite{jie2023fact} tensorizes the weights of each ViT into a single 3D tensor and updates it during finetuning.
Apart from additional modules in the network, Visual prompting~\cite{bahng2022visual} proposes learning tunable parameters in the input space.
MAM-Adapter~\cite{hetowards} formulates these works in a unified framework, and NOAH~\cite{zhang2022neural} searches for the optimal design of prompt modules for downstream tasks.
Apart from tuning a pre-trained Vision Transformer, CoOp~\cite{zhou2022learning} and CoCoOp~\cite{zhou2022conditional} explores the application of prompt tuning for CLIP~\cite{radford2021learning} via learning textual prompts. Maple~\cite{khattak2023maple} further promotes strong coupling between the vision-language prompts to ensure mutual synergy.

\section{From Old Classes to New Classes}

\subsection{Class-Incremental Learning}

CIL aims to learn from an evolving data stream with new classes to build a unified classifier~\cite{rebuffi2017icarl}. There is a sequence of $B$ training tasks $\left\{\D^{1}, \D^{2}, \cdots, \D^{B}\right\}$, where $\D^{b}=\left\{\left(\x_{i}^{b}, y_{i}^{b}\right)\right\}_{i=1}^{n_b}$ is the $b$-th incremental step with $n_b$ instances. Here, the training instance $\x_i^b \in \R^D$ belongs to class $y_i \in Y_b$, where $Y_b$ is the label space of task $b$. $Y_b  \cap Y_{b^\prime} = \varnothing$ for $b\neq b^\prime$. During the $b$-th training stage, we can only access data from $\D^b$ for model updating.
This paper focuses on the {\bf exemplar-free} CIL setting~\cite{zhu2021prototype,wang2022learning}, where {\em no historical data} can be fetched for rehearsal.
The goal of CIL is to build a unified model for all seen classes incrementally, \ie, acquiring knowledge from new classes and meanwhile preserving knowledge from former ones. 
The model's capability is evaluated over all seen classes $\mathcal{Y}_b=Y_1 \cup \cdots Y_b$ after each incremental task. Formally, the target is to fit a model $f(\x): X\rightarrow\mathcal{Y}_b$ that minimizes the empirical risk across all testing datasets:
\begin{equation} \label{eq:cilrisk} 
	\textstyle
	\frac{1}{N}\sum_{(\mathbf{x}_j, y_j) \in \mathcal{D}_{t}^1\cup\cdots\mathcal{D}_{t}^b} \ell \left(f
	\left(\x_{j}	\right), {y}_{j}\right) \,,
\end{equation}
where $\ell(\cdot,\cdot)$ measures the discrepancy between prediction and ground-truth label.  $\mathcal{D}_{t}^b$ denotes the testing set of task $b$, and $N$ is the number of instances. A good CIL model satisfying Eq.~\ref{eq:cilrisk} has discriminability among all classes, which strikes a balance between learning new classes and remembering old ones.

Following \cite{wang2022learning,wang2022dualprompt,zhou2024continual}, we assume the availability of a pre-trained model (\eg, a ViT~\cite{dosovitskiy2020image} or ResNet~\cite{he2016deep}) on ImageNet~\cite{deng2009imagenet}, which we use as the initialization of $f(\x)$. For clarity, we decouple the deep model into two parts: $f(\x)=W^{\top}\phi(\x)$, where $\phi(\cdot):\mathbb{R}^{D} \rightarrow \mathbb{R}^{d}$ is the embedding function and $W\in\mathbb{R}^{d\times |\mathcal{Y}_{b}|}$ is the classification head. We denote the classifier for class $k$ as $\w_k$: $W=[\w_1,\cdots,\w_{|\mathcal{Y}_{b}|}]$. 
We refer to the features after pooling as $\phi(\x)$ for convolutional networks.
In a plain ViT, the input encoding layer transforms the image into a sequence of output features $\x_e\in\R^{L\times d}$, where $L$ is the sequence length. We assume the first token in $\x_e$ to be the $\texttt{[CLS]}$ token to simplify notation. 
$\x_e$ is then fed into the subsequent layers (\ie, multi-head self-attention and MLP) to produce the final embeddings. 
We treat the embedded $\texttt{[CLS]}$ token as $\phi(\x)$ for ViT.

\begin{figure}[t]
	\centering
	{
		\subfigure[Accuracy of new classes]
		{\includegraphics[width=.469\columnwidth]{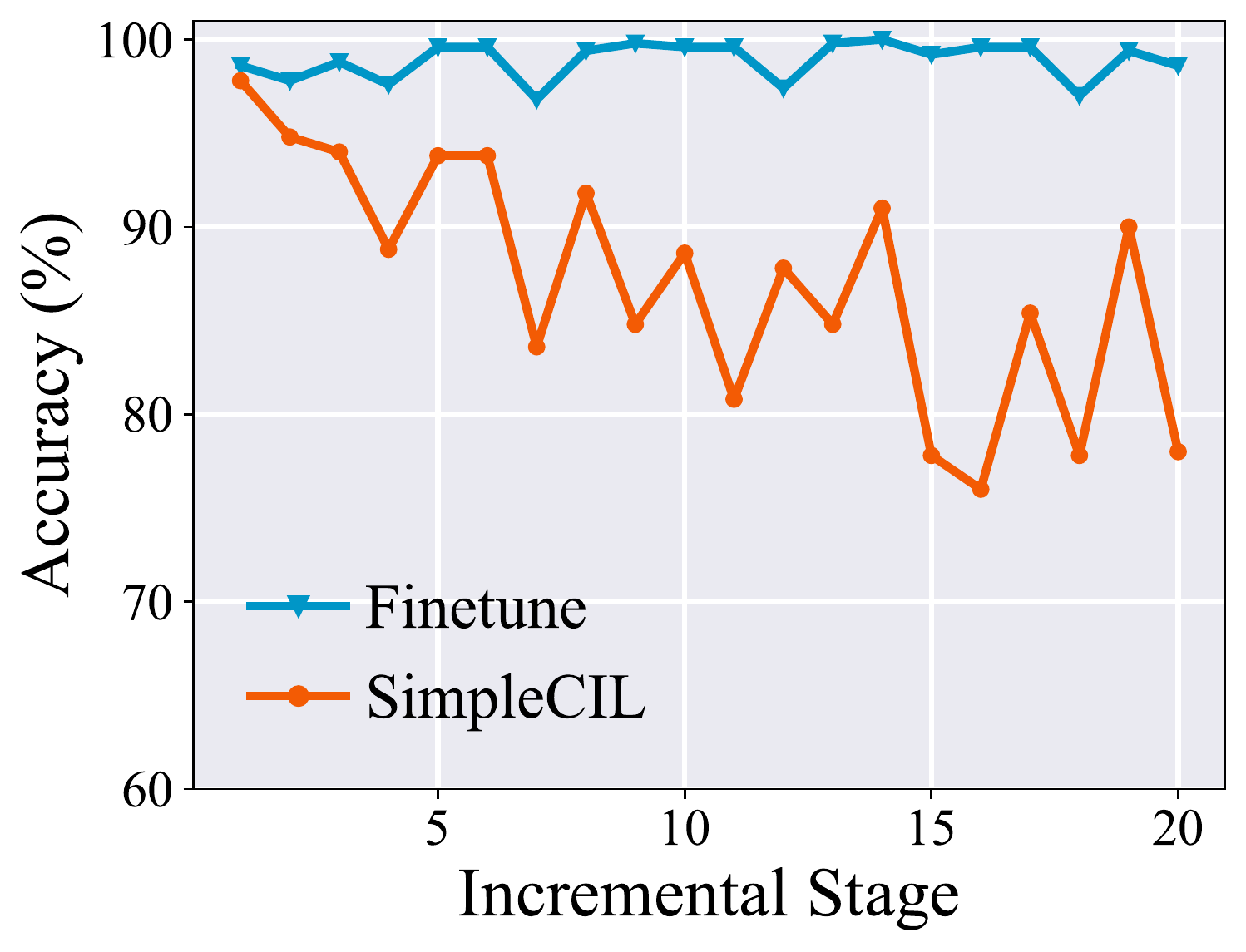}
			\label{figure:new-perf}
		}
		\hfill
		\subfigure[Accuracy of old classes]
		{\includegraphics[width=.469\columnwidth]{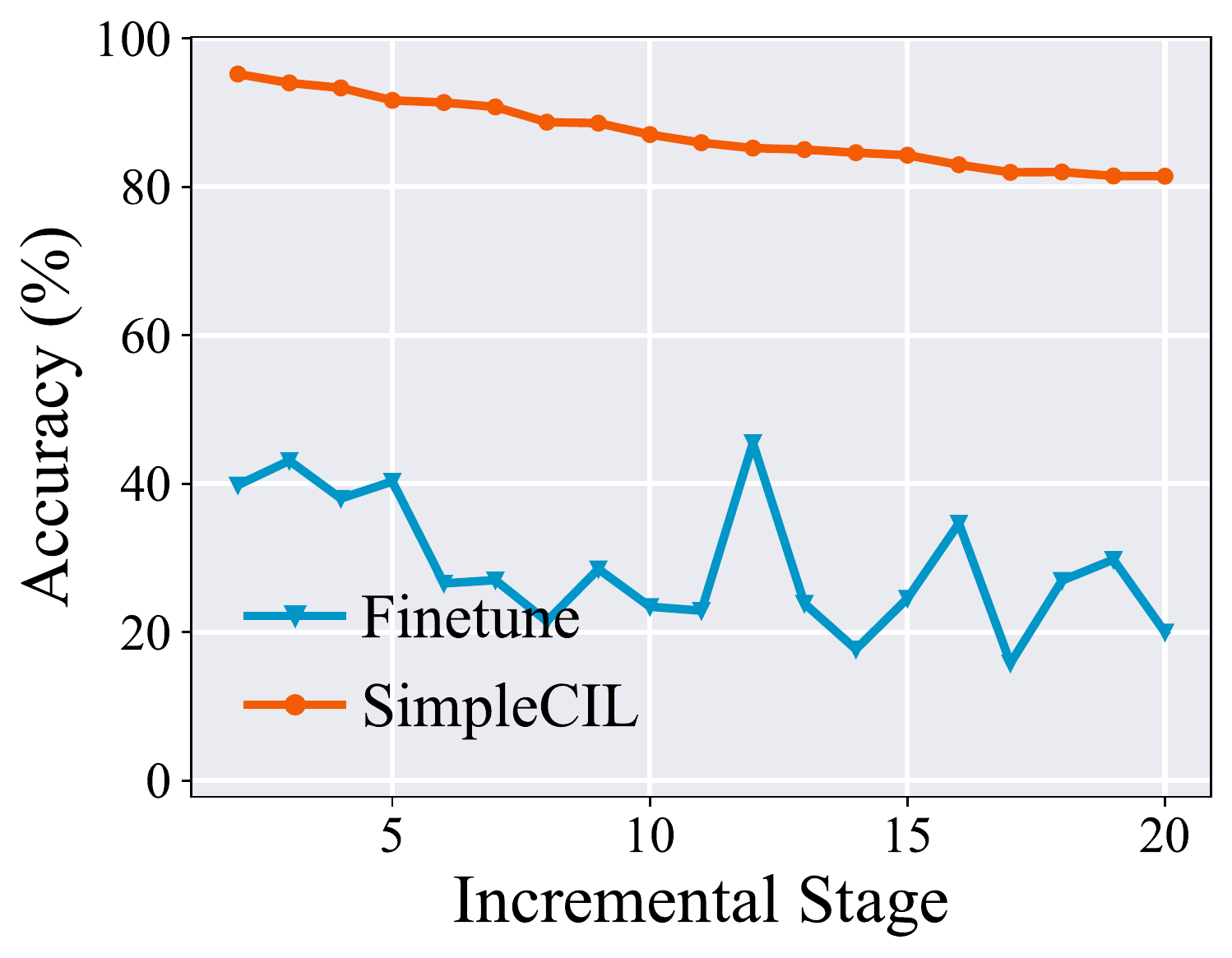}
			\label{figure:old-perf}
		}	
	}
	\caption{\small Performance of new and old classes in CIL with PTM. Sequentially finetuning the model fills the domain gap and performs better on new classes, while freezing the model has better generalizability and performs better on old classes.
	} 
	\label{figure:old_and_new_perf}
\end{figure}

\subsection{Adaptivity and Generalizability in Class-Incremental Learning}

\noindent\textbf{CIL with Adaptivity:} Before introducing PTMs into CIL, models are trained from scratch to gradually acquire knowledge of new classes. The common solution is to update the incremental model with cross-entropy loss, which equips the model with {\em adaptivity} to adapt to new tasks: 
\begin{equation} \label{eq:adaptivity} \textstyle
	\mathcal{L}=	\sum_{(\mathbf{x}_i, y_i) \in \mathcal{D}^b} \ell \left(f
	\left(\x_{i}	\right) , {y}_{i}\right)+ \mathcal{L}_{reg} \,,
\end{equation}
where $\mathcal{L}_{reg}$ stands for the regularization terms to resist forgetting, \eg, knowledge distillation~\cite{hinton2015distilling,li2017learning} or parameter regularization~\cite{kirkpatrick2017overcoming}.

\noindent\textbf{CIL with Generalizability:} With the introduction of PTM to CIL~\cite{wang2022learning}, continual learners are born with {\em generalizability}, which can be directly transferred to downstream tasks without learning. 
Correspondingly, we define a simple baseline, {\bf SimpleCIL}, to transfer PTM for incremental tasks.
With the embedding function $\phi(\cdot)$ {\em frozen} throughout the learning process, we extract average embedding (\ie, prototype~\cite{snell2017prototypical}) of each class:
\begin{align} \label{eq:prototype}\textstyle
	\p_i=\frac{1}{K}{\sum_{j=1}^{|\mathcal{{D}}^b|}\mathbb{I}(y_j=i)\phi(\x_j)}
	\,,
\end{align}
where $K={\sum_{j=1}^{|\mathcal{{D}}^b|}\mathbb{I}(y_j=i)}$, and $\mathbb{I}(\cdot)$ is the indicator function. The averaged embedding represents the most common pattern of the corresponding class. We set the prototype as the classifier, \ie, $\w_i=\p_i$, to directly adjust the PTM for CIL. SimpleCIL demonstrates competitive performance in Figure~\ref{figure:intro}, confirming the strong generalizability of pre-trained models.

\noindent\textbf{Generalizability vs. Adaptivity:}
Eq.~\ref{eq:adaptivity} and Eq.~\ref{eq:prototype} address different aspects of CIL models. The former aims to enhance the adaptivity by enabling the model to be gradually tuned. By contrast, the latter highlights the model's generalizability by freezing it throughout the learning process. 
To understand their roles in CIL, we conduct an experiment on CIFAR100 with 20 incremental tasks and compare the performance of finetuning versus SimpleCIL. These methods are based on pre-trained ViT-B/16-IN21K, and we separately report the performance of new ($Y_b$) and old ($\mathcal{Y}_{b-1}$) classes in Figure~\ref{figure:old_and_new_perf}. Specifically, SimpleCIL relies on the generalizability of PTM, which works competitively even without training on the target dataset. However, it can be further improved to grasp the task-specific features, and finetuning shows better performance in new classes with the help of adaptivity. However, finetuning suffers catastrophic forgetting of old classes since features are continually changing. 

To summarize, these characteristics are two core aspects of CIL --- adaptivity enables the model to bridge the domain gap between pre-training and incremental learning. At the same time, generalizability encourages knowledge transfer from pre-training to incremental learning. Therefore, both of them should be cultivated to facilitate CIL.

\begin{figure*}[t]
	\begin{center}
		\includegraphics[width=1.98\columnwidth]{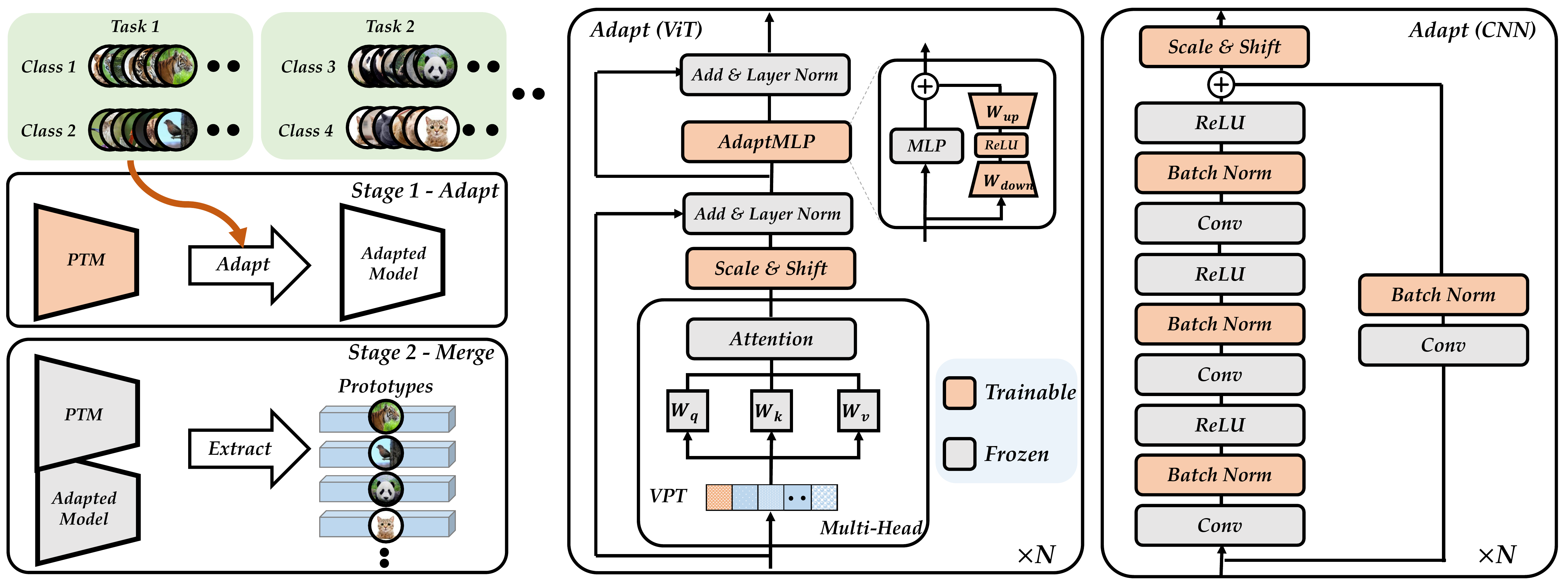}
	\end{center}
	\caption{\small  Illustration of \mame. {\bf Left}: the training protocol of \mame. We adapt the PTM using the first stage training set $\D^1$ and then concatenate the embedding functions of PTM and the adapted model to maintain {\em generalizability} and {\em adaptivity}. The aggregated embedding function $[\phi^*(\cdot),\phi(\cdot)]$ is frozen throughout the following stages, and we extract the prototypes via Eq.~\ref{eq:prototype-ours} to set the classifier.
		{\bf Middle}: adapting pre-trained ViT for CIL. We provide VPT Deep/Shallow, Scale \& Shift, and Adapter for model adaptation.
		{\bf Right}: adapting pre-trained CNN for CIL. We provide BN tuning and Scale \& Shift for model adaptation. \name is a general framework that can be orthogonally combined with these adapting techniques. Red modules in the figure are trainable, while gray ones are frozen.
	}
	\label{figure:teaser}
\end{figure*} 

\section{{\scshape{Aper}}: AdaPt and mERge PTMs for CIL}

Motivated by the potential for enhancing generalizability and adaptivity, can we achieve these characteristics in a unified framework? Specifically, we aim to achieve this goal from two aspects. On the one hand, to bridge the domain gap between the PTM and downstream datasets, {\em model adaptation} is essential to move the PTM towards incremental data. On the other hand, since the adapted model may lose the generalizability of high-level features, we attempt to {\em merge} the adapted model and PTM into a unified network for future tasks. The merged embedding function is kept frozen throughout the incremental learning process, transferring the generalizable embedding of model sets to incoming new classes. In this way, generalizability and adaptivity are achieved in the unified framework.
We first introduce the framework of \name and then discuss the specific techniques for model adaptation.

\subsection{Training Procedure of \name}

Although PTMs have discriminating features, there may exist a significant domain gap between the pre-trained dataset and incremental data. For example, the PTM is optimized to capture the characteristics of classes in ImageNet, 
while the incremental data stream may correspond to specialized data requiring domain knowledge or has extensive concept drift from ImageNet.
To bridge this gap, an adapting process can be developed with the incremental data:
\begin{align} \label{eq:adapt}
	f^*(\x)=\mathcal{F}(f(\x),\D,\Theta) \,,
\end{align}
where the adapting algorithm $\mathcal{F}$ takes the current model $f(\x)$ and the dataset $\D$ as input. It optimizes the parameter set $\Theta$ and produces the adapted model $f^*(\x)$ that gains the domain-specific knowledge in the corresponding dataset. We introduce the variations of $\mathcal{F}$ in Section~\ref{sec:adaptation}.
If we could obtain all the incremental training sets at once, adapting the model via $\mathcal{F}(f(\x),\D^{1}\cup \D^{2} \cdots\cup \D^{B},\Theta)$ can transfer the knowledge from the PTM to the incremental dataset and grasp the task-specific features for better performance.

However, since data in CIL arrive sequentially, we cannot hold all the training sets at once. Continuously adapting the model would consequently result in catastrophic forgetting (as shown in Figure~\ref{figure:old-perf}). Hence, an alternative choice is to adapt the model only {\em in the first incremental stage}:
\begin{align} \label{eq:adapt_ours}
	f^*(\x)=\mathcal{F}(f(\x),\D^1,\Theta) \,.
\end{align}
Since $\D^1$ is a subset of the incremental data stream, it also possesses {\em domain-specific} knowledge that could facilitate model adaptation.
The tuning process enhances the adaptivity of the CIL model, and the next question is to ensure {\em generalizability}. Since  Eq.~\ref{eq:adapt_ours} forces the original generalizable feature to become more specialized to the downstream task, high-level features irrelevant to $\D^1$ shall be {\em overwritten and forgotten}. Therefore, a better solution is to concatenate the features extracted by the PTM and the adapted model, \ie, $[\phi^*(\x),\phi(\x)]$, where $\phi^*(\x)$ and $\phi(\x)$ stand for the adapted and pre-trained embedding functions, respectively.

To maintain generalizability, we {\em freeze} the concatenated embedding functions $[\phi^*(\cdot),\phi(\cdot)]$ after adaptation and extract prototypes for the following classes:
\begin{align} \label{eq:prototype-ours}
	\textstyle
	\p_i=\frac{1}{K}
	{\sum_{j=1}^{|\mathcal{{D}}^b|}\mathbb{I}(y_j=i)[\phi^*(\x_j),\phi(\x_j)]}
	\,,
\end{align}
where $K={\sum_{j=1}^{|\mathcal{{D}}^b|}\mathbb{I}(y_j=i)}$. Compared to Eq.~\ref{eq:prototype}, Eq.~\ref{eq:prototype-ours} contains additional information from the adapted model, which incorporates domain-specific features for better recognition. These prototypes reveal the most common patterns from the adapted and pre-trained models, ensuring both generalizability and adaptivity. We directly adopt the class prototype as the classifier weight, \ie, $\w_i=\p_i$, and utilize a cosine classifier for classification: $f(\mathbf{x})=(\frac{W}{\|W\|_{2}})^{\top}(\frac{[\phi^*(\x),\phi(\x)]}{\|[\phi^*(\x),\phi(\x)]\|_{2}})$. Based on the similarity between instance embedding and class prototype, it assigns a higher probability to the class with a more similar prototype.

\noindent\bfname{Effect of Adapt and Merge:} We give the visualizations of \name in Figure~\ref{figure:teaser} (left). Although $\D^1$ is a subset of the entire training set, adapting with it still helps transfer the PTM from the upstream dataset to the downstream task. The adapting process can be viewed as a further pre-training procedure, which adapts the PTM to the incremental dataset and bridges the domain gap. By merging the embedding functions of the PTM and the adapted model, the extracted features are more representative than any one of them alone. Additionally, since the model is only trainable in the first incremental task, the efficiency of \name is comparable to SimpleCIL, which does not require sequential tuning. On the other hand, since the model is frozen in the subsequent tasks, it does not suffer catastrophic forgetting of former concepts. We give the pseudo-code of \name in Algorithm~\ref{alg1}. 
Given the pre-trained model, we first adapt it with the first training dataset via Eq.~\ref{eq:adapt_ours} to get the adapted model.
Afterward, we freeze the pre-trained model and adapted model and merge the embeddings. For the subsequent tasks, we get a new dataset and replace the classifier weights with prototypical features (\ie, class centers).
In the extreme case where the adaptation process in Eq.~\ref{eq:adapt_ours} does nothing to the PTM, \name will degrade to SimpleCIL, which guarantees the performance lower bound.

\begin{algorithm}[t]
	\caption{AdaPt and mERge (\mame) for CIL }
	\label{alg1}
	\raggedright
	{\bf Input}: Incremental datasets: $\left\{\D^{1}, \D^{2}, \cdots, \D^{B}\right\}$, Pre-trained Model: $f(\x)$;\\
	{\bf Output}: Updated model; 
	\begin{algorithmic}[1]
			\State Adapt the model to $\D^1$ via Eq.~\ref{eq:adapt_ours}; \label{line:1}  {\Comment{\color{cyan}{Model adapt}}}
			\State Freeze the embedding functions $\phi^*(\cdot)$ and $\phi(\cdot)$;
			\State Merge the embeddings, \ie,  $[\phi^*(\x),\phi(\x)]$;{\Comment{\color{cyan}{Model merge}}}
			\For{$b=1,2\cdots,B$}  {\Comment{\color{cyan}{Incremental learning}}}
			\State Get the incremental training set $\D^b$; 
			\State Extract the prototypes via Eq.~\ref{eq:prototype-ours}; 
			\State Replace the classifier with prototype; \label{line:7}
			\EndFor 
			\Return the updated model;
	\end{algorithmic}
\end{algorithm}

\subsection{Adapting the PTM} \label{sec:adaptation}

To bridge the distribution gap between the pre-trained and incremental datasets, \mame's performance depends on the effective adapting algorithm $\mathcal{F}$.
In this section, we discuss six specializations of $\mathcal{F}$ in \name that can handle different types of PTMs, such as ViTs and CNNs.

\noindent\textbf{Fully Finetune}: is a common solution when transferring the model to downstream tasks. It involves tuning all parameters in the adapting process, \ie, $\Theta=\theta_{\phi}\cup\theta_{W}$, and minimizing the discrepancy between the model's output and the ground truth:
\begin{align} \label{eq:adapt-ft}
	\min_{\theta_{\phi}\cup\theta_{W}} \sum_{(\mathbf{x}_j, y_j) \in \mathcal{D}^1} \ell \left(f
	\left(\x_{j}	\right), {y}_{j}\right) \,.
\end{align}
However, the tuning cost could be relatively {\em high} for large-scale PTMs, \eg, ViTs. Therefore, some parameter-efficient tuning techniques can alleviate the tuning cost and be better solutions.

\noindent\textbf{Visual Prompt Tuning (VPT)~\cite{jia2022visual}}: is a lightweight tuning technique for adapting ViTs, which only prepends some learnable prompts $\mathbf{P}\in\R^{p\times d}$ to form the extended features $[\mathbf{P},\x_e]$, where $\x_e$ is the encoded features of the input image. The extended features are then fed into the subsequent layers of ViT to calculate the final embeddings. There are two variations of VPT: \textbf{VPT-Deep}, which prepends the prompts at every attention layer, and \textbf{VPT-Shallow}, which only prepends the prompts at the first layer. During optimization, it freezes the pre-trained weights in the embedding function and optimizes these prompts and classification head, \ie, $\Theta=\theta_{\mathbf{P}}\cup\theta_{W}$.

\noindent\textbf{Scale \& Shift (SSF)~\cite{lianscaling}}: aims to adjust the feature activation by scaling and shifting. It  appends an extra SSF layer after each operation layer (\ie, MSA and MLP) and adjusts the output of these operations. Given the input $\x_i\in\R^{L\times d}$, the output $\x_o\in\R^{L\times d}$ is formulated as:
\begin{align} \label{eq:adapt-ssf}	
	\textstyle
	\x_o=\gamma \otimes \x_i+\beta \,,
\end{align}
where $\gamma\in\R^d$ and $\beta\in\R^d$ are the scale and shift factors, respectively. $\otimes$ is Hadamard product (element-wise multiplication). The model optimizes the SSF layers and classifier, \ie, $\Theta=\theta_{SSF}\cup\theta_{W}$, to trace the features of new tasks.

\noindent\textbf{Adapter~\cite{houlsby2019parameter,chenadaptformer}}: is a bottleneck module which contains a down-projection $W_\text{down}\in\R^{d\times r}$ to reduce the feature dimension, a non-linear activation function, and an up-projection $W_\text{up}\in\R^{r\times d}$ to project back to the original dimension. We follow~\cite{chenadaptformer} to equip the original MLP structure in ViT with the adapter. We denote the input of the MLP layer as $\x_\ell$, and the output of AdaptMLP is formatted as:
\begin{align} \label{eq:adapt-adapter}
	\text{MLP}(\x_\ell)+\text{ReLU}(\x_\ell W_\text{down})W_\text{up} \,.
\end{align}
With pre-trained weights frozen, it optimizes the adapter and classification head, \ie, $\Theta=\theta_{W_\text{down}}\cup\theta_{W_\text{up}}  \cup \theta_{W}$.\\
\noindent\textbf{Batch Normalization Tuning}: If the PTM is a convolutional network, \eg, CNNs, we can adjust the BN~\cite{ioffe2015batch} parameters. Since the running mean and variance in BN are compatible with the upstream data distribution, they could be {\em unstable} for downstream tasks. Correspondingly, we can reset the running statistics in BN and adapt to the current data via forward passing. No backpropagation is required, making it quick and simple for the pre-trained model.

\noindent\bfname{Discussions:} We visualize the adapting process of \name in Figure~\ref{figure:teaser}.
Compared to fully finetuning, parameter-efficient tuning adjusts the PTM towards the downstream task with lightweight modules. Besides, recent parameter-efficient tuning methods show stronger performance than fully finetune, indicating stronger adaptivity in the adapting process.
The adapted model can capture the specialized features in the incremental data, leading to better adaptivity. Since L2P and DualPrompt are based on pre-trained ViT, they cannot be deployed with CNN. In contrast, \name is a general framework that efficiently handles diverse structures. Specifically, \name can be combined with VPT/SSF/Adapter for ViT and SSF/BN Tuning for CNN.
Since \name adopts the prototype-based classifier, the linear classifier $W$ will be dropped after adaptation.

\begin{table*}[t]
	\caption{\small Average and last performance comparison on seven datasets with {\bf ViT-B/16-IN21K} as the backbone.  `IN-R/A' stands for `ImageNet-R/A,' `ObjNet' stands for `ObjectNet,' and `OmniBench' stands for `OmniBenchmark.' 
		The best performance is shown in bold.
	}
	\label{tab:benchmark}
	\centering
	\resizebox{1.0\textwidth}{!}{%
		\begin{tabular}{@{}lccccccccc cccccccc}
			\toprule
			\multicolumn{1}{l}{\multirow{2}{*}{Method}} & 
			\multicolumn{2}{c}{CIFAR B0 Inc5} & \multicolumn{2}{c}{CUB B0 Inc10} 
			& \multicolumn{2}{c}{IN-R B0 Inc5}
			& \multicolumn{2}{c}{IN-A B0 Inc10}
			& \multicolumn{2}{c}{ObjNet B0 Inc10}
			& \multicolumn{2}{c}{OmniBench B0 Inc30}
			& \multicolumn{2}{c}{VTAB B0 Inc10} \\
			& {$\bar{\mathcal{A}}$} & ${\mathcal{A}_B}$  
			& {$\bar{\mathcal{A}}$} & ${\mathcal{A}_B}$
			& {$\bar{\mathcal{A}}$} & ${\mathcal{A}_B}$   
			& {$\bar{\mathcal{A}}$} & ${\mathcal{A}_B}$ 
			& {$\bar{\mathcal{A}}$} & ${\mathcal{A}_B}$ 
			& {$\bar{\mathcal{A}}$} & ${\mathcal{A}_B}$ 
			& {$\bar{\mathcal{A}}$} & ${\mathcal{A}_B}$ 
			\\
			\midrule
			Finetune	& 38.90 & 20.17 &26.08 & 13.96 &21.61 & 10.79 &21.60 & 10.96 & 19.14 & 8.73 & 23.61 & 10.57 & 34.95 & 21.25  \\
			Finetune Adapter~\cite{chenadaptformer} & 60.51 &49.32& 66.84 &52.99 & 47.59 &40.28 &43.05 &37.66 &50.22 &35.95 &62.32& 50.53 &48.91 & 45.12 \\
			LwF~\cite{li2017learning}& 46.29 & 41.07 &48.97 & 32.03  & 39.93 &26.47 & 35.39 &23.83 & 33.01 & 20.65 & 47.14 &33.95 & 40.48 & 27.54\\
			SDC~\cite{yu2020semantic} & 68.21 &63.05 &  70.62 & 66.37 & 52.17 & 49.20 & 26.65 & 23.57 & 39.04 & 29.06 &60.94 & 50.28 &45.06 &22.50\\
			L2P~\cite{wang2022learning}   & 85.94 & 79.93 &67.05 & 56.25 & 66.53 & 59.22 & 47.16 & 38.48 &  63.78 & 52.19 &73.36 & 64.69 & 77.11 & 77.10\\
			DualPrompt~\cite{wang2022dualprompt}   &87.87 & 81.15& 77.47 & 66.54 & 63.31 & 55.22 & 52.56 & 42.68 & 59.27 & 49.33 & 73.92 & 65.52 & 83.36 & 81.23\\
			CODA-Prompt~\cite{smith2023coda} & 89.11 & 81.96 & 84.00 & 73.37 & 64.42 &55.08 & 48.51 & 36.47 & 66.07 &53.29 &77.03 &68.09 &83.90 &83.02\\
			CPP~\cite{li2024steering} & 85.21 & 78.64 & 86.60 & 85.27 & 64.33 & 60.74 & 53.70 & 40.70 &  60.44 & 49.92 &  71.52 & 73.26 & 85.92 &  84.30 \\
			LAE~\cite{gao2023unified} & \bf 92.47 & \bf 87.62 & 83.13 & 77.78 & 69.05 & 63.17 & 57.19 & 46.41 & 62.28 & 50.57 & 73.80 & 70.63 &  86.14 & 84.39 \\
			\midrule
			SimpleCIL   &  87.57 & 81.26 & 92.20 &\bf 86.73 & 62.58 & 54.55 & 60.50 & 49.44 & 65.45 & 53.59 & 79.34 & 73.15 & 85.99 & 84.38\\
			\rowcolor{LightCyan}\name w/ Finetune   & 87.67 & 81.27 & 91.82 & 86.39 & 70.51 & 62.42 &   61.57 & 50.76 & 61.41 & 48.34 & 73.02 & 65.03 &  \bf 87.47 & 80.44\\
			\rowcolor{LightCyan}\name w/ VPT-Shallow    &  90.43 & 84.57& 92.02 &86.51 & 66.63 &58.32 & 57.72 &46.15 & 64.54 & 52.53 & 79.63 & 73.68 & 87.15 &\bf 85.36\\
			\rowcolor{LightCyan}\name w/ VPT-Deep & 88.46 & 82.17 & 91.02 &84.99 & 68.79 & 60.48 & 60.59 & 48.72 & 67.83 & 54.65 & \bf 81.05 & \bf 74.47 & 86.59 & 83.06\\
			\rowcolor{LightCyan}\name w/ SSF  & 87.78 & 81.98   & 91.72 &86.13& 68.94 & 60.60 & \bf 62.81 &  \bf 51.48 & \bf69.15 &\bf 56.64 &  80.53 & 74.00 & 85.66 & 81.92\\
			\rowcolor{LightCyan}\name w/ Adapter &   90.65 & 85.15 &\bf92.21 &\bf86.73 &\bf 72.35 &\bf 64.33 & 60.53 & 49.57 &  67.18 & 55.24 &  80.75 & 74.37 &  85.95 & 84.35\\
			\bottomrule
		\end{tabular}
	}
\end{table*}

\subsection{Discussions on related concepts}
There is a famous concept in CIL, namely ``stability-plasticity dilemma''~\cite{grossberg2012studies,mermillod2013stability,mirzadeh2020understanding}. Specifically, ``stability'' denotes the ability of a continual learner to remember old knowledge, while ``plasticity'' refers to the ability to learn new concepts. These concepts are similar to the concept of ``generalizability and adaptivity'' in this paper. However, there are some main differences that need to be highlighted.\\
Firstly, ``stability-plasticity dilemma'' mainly refers to the problem of {\em training from scratch} (\ie, randomly initialized weights), where the model needs to balance learning new concepts and remembering the old. These concepts are two ultimate goals of continual learning, which do not conflict with ``generalizability and adaptivity'' raised in this paper.
Secondly, ``Generalizability and adaptivity'' in this paper is the new characteristic in the era of PTMs. Specifically, a randomly initialized model does not have such ``generalizability,'' which cannot be directly applied to the downstream tasks. However, continual learners are born with ``generalizability'' if starting with a PTM, and we observe a simple baseline shows strong performance. Furthermore, we find that ``generalizability'' is insufficient for all downstream tasks, especially when downstream tasks come from a different distribution. In this case, we need to enhance the PTM's ``adaptivity'' by further tuning it with the downstream task. Finally, by aggregating the features extracted by the pre-trained and adapted models, we unify these characteristics in a single model.

In summary, the ``generalizability and adaptivity'' in this paper is a new characteristic in class-incremental learning with pre-trained models. We aim to unify these characteristics in CIL and propose our \name by aggregating the adapted and pre-trained models.

\begin{figure*}[t]
	\begin{center}
		\subfigure[CIFAR B50 Inc5]
		{\includegraphics[width=.65\columnwidth]{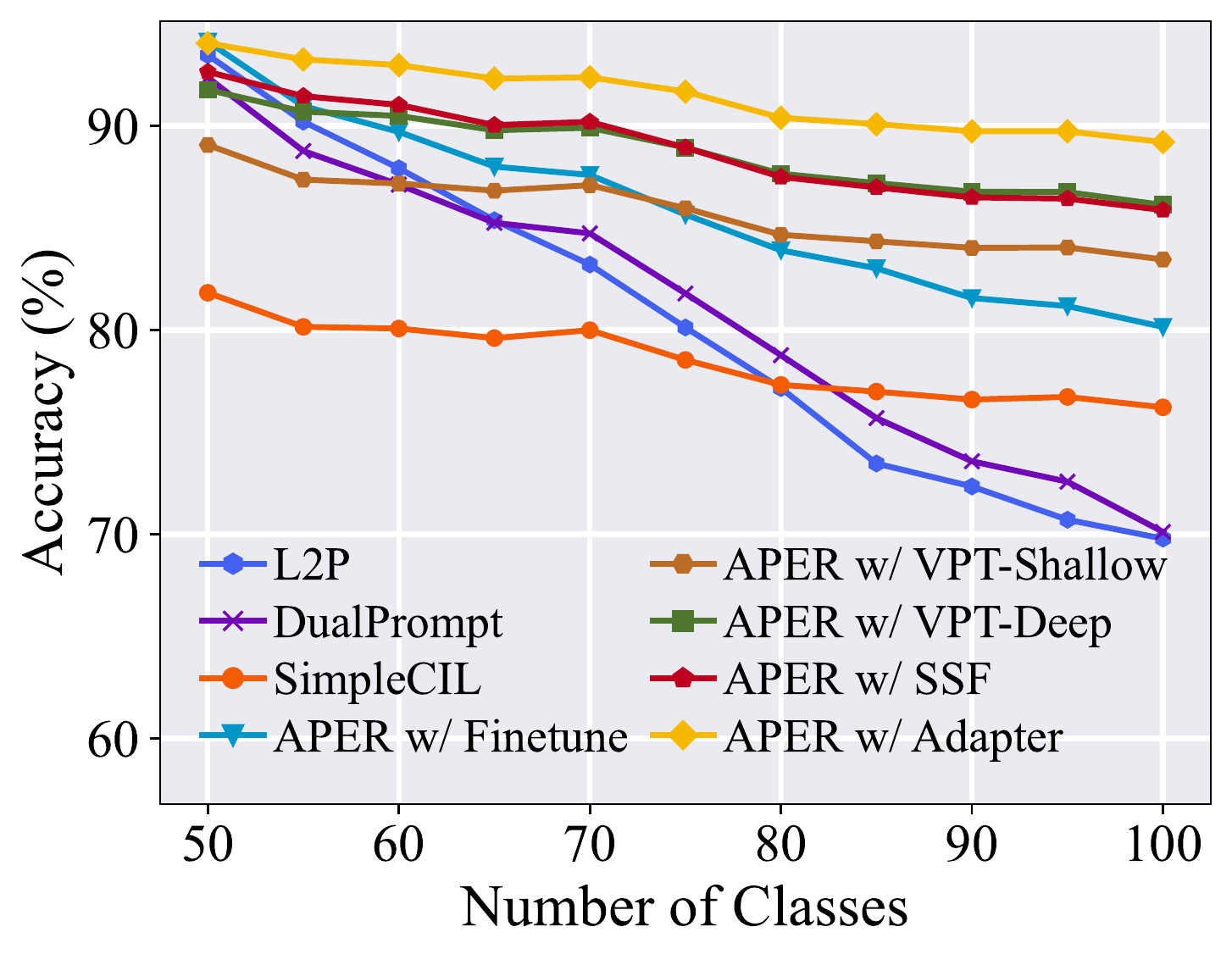}
			\label{figure:vit-cifar}
		}
		\subfigure[CUB B100 Inc5]
		{\includegraphics[width=.65\columnwidth]{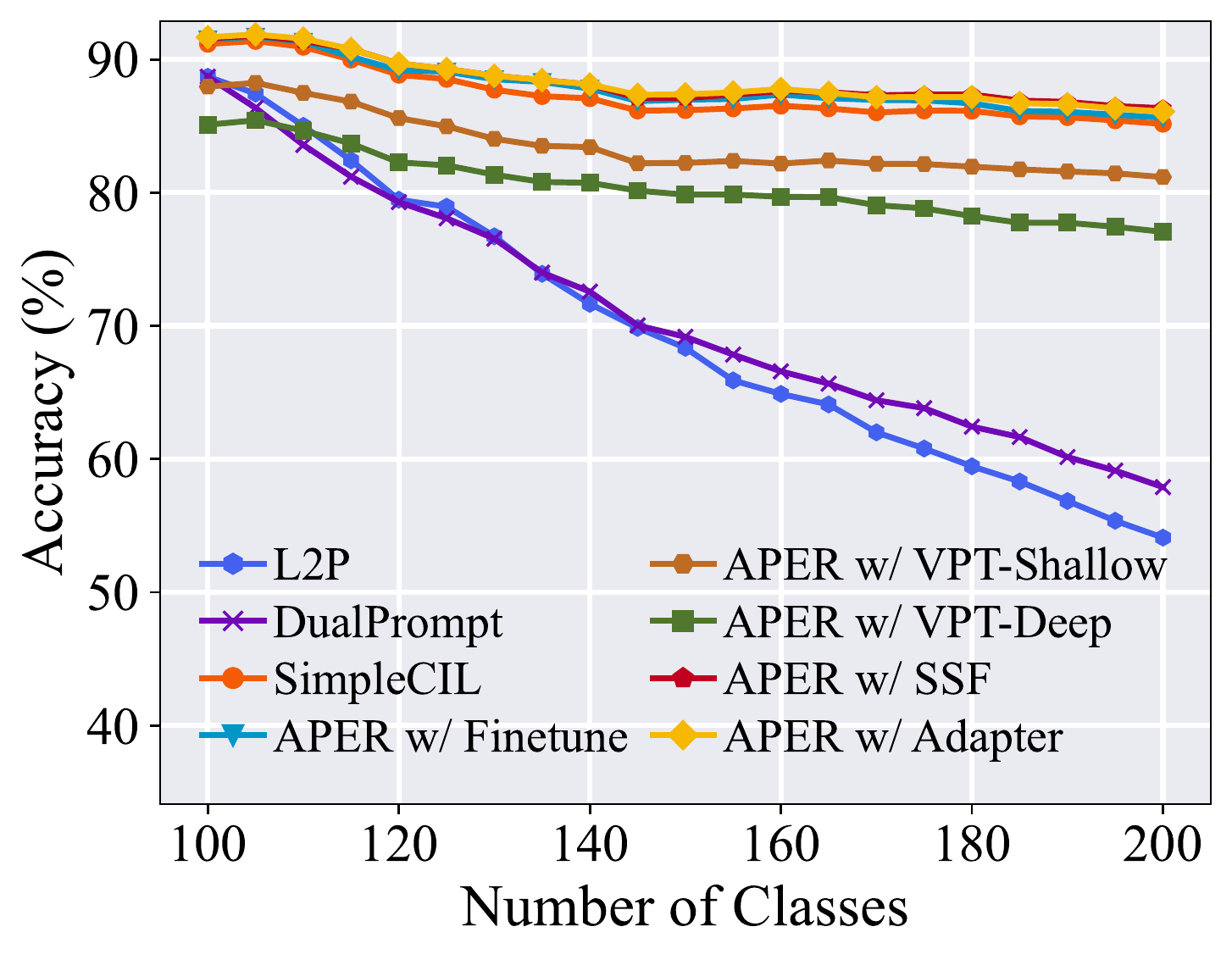}
			\label{figure:vit-cub}
		}
		\subfigure[ImageNet-A B100 Inc5]
		{\includegraphics[width=.65\columnwidth]{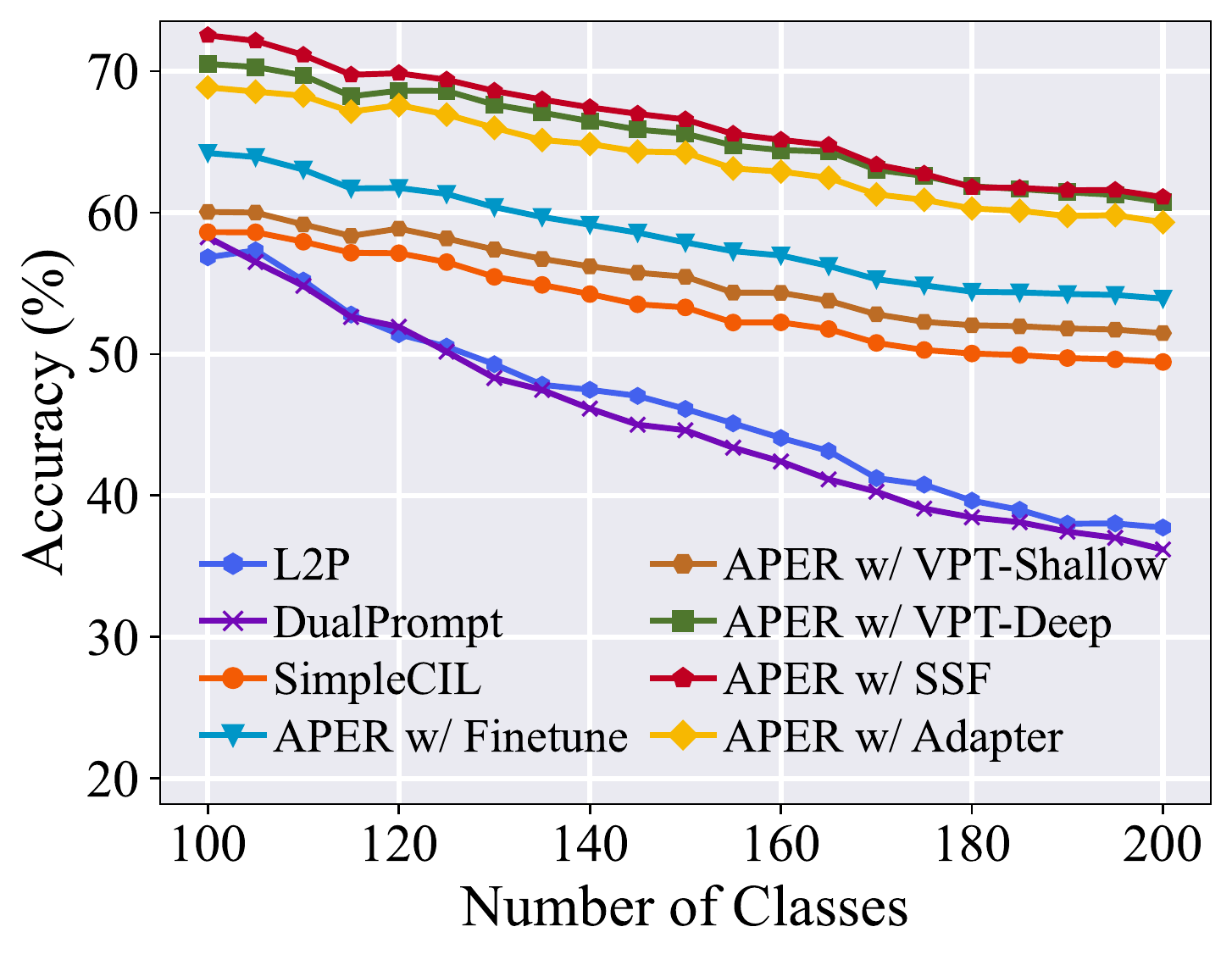}
			\label{figure:vit-ina}
		}	
		\subfigure[ImageNet-R B100 Inc5]
		{\includegraphics[width=.65\columnwidth]{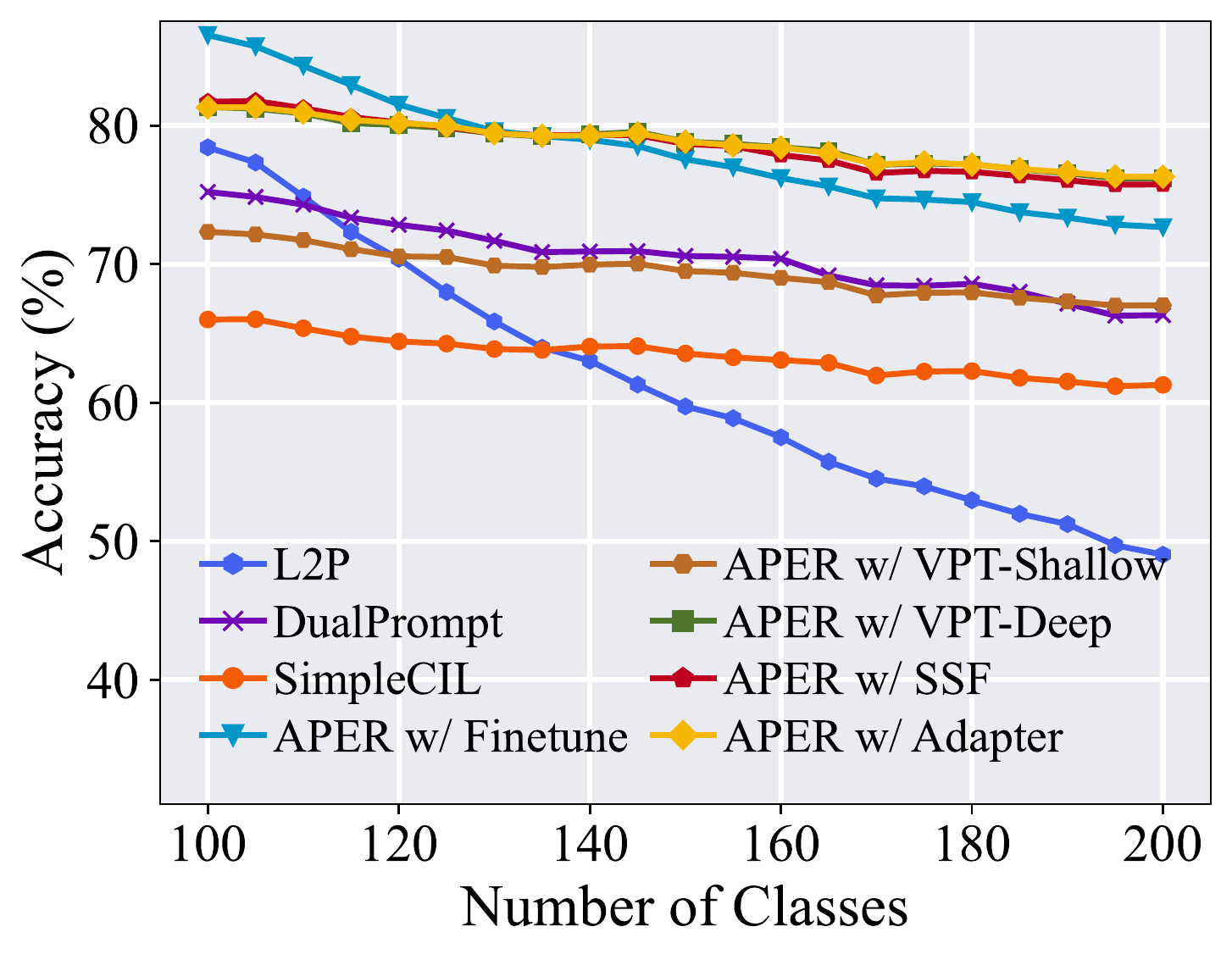}
			\label{figure:vit-inr}
		}	
		\subfigure[ OmniBenchark B150 Inc15]
		{\includegraphics[width=.65\columnwidth]{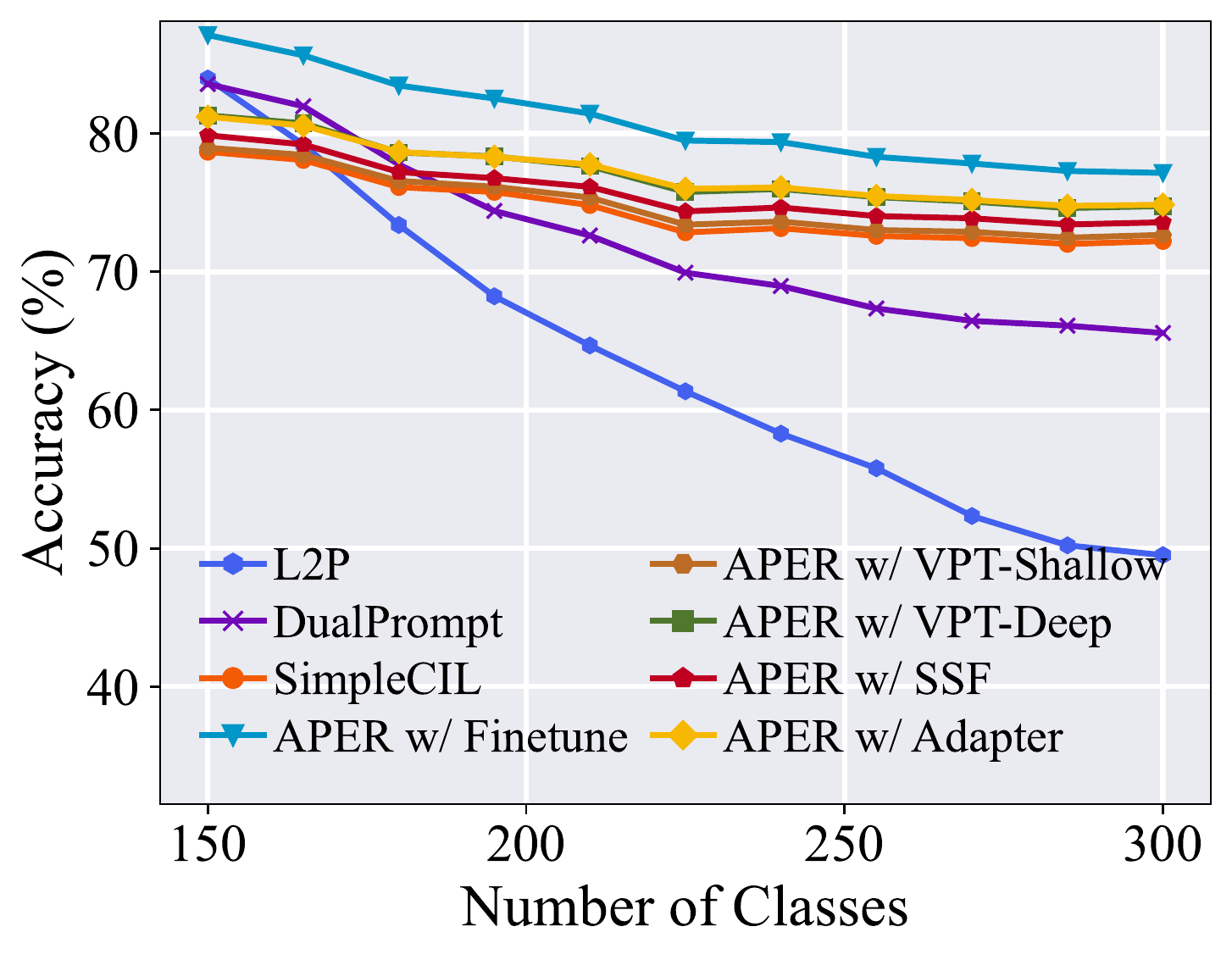}
			\label{figure:vit-omni}}
		\subfigure[ObjectNet B100 Inc5]
		{\includegraphics[width=.65\columnwidth]{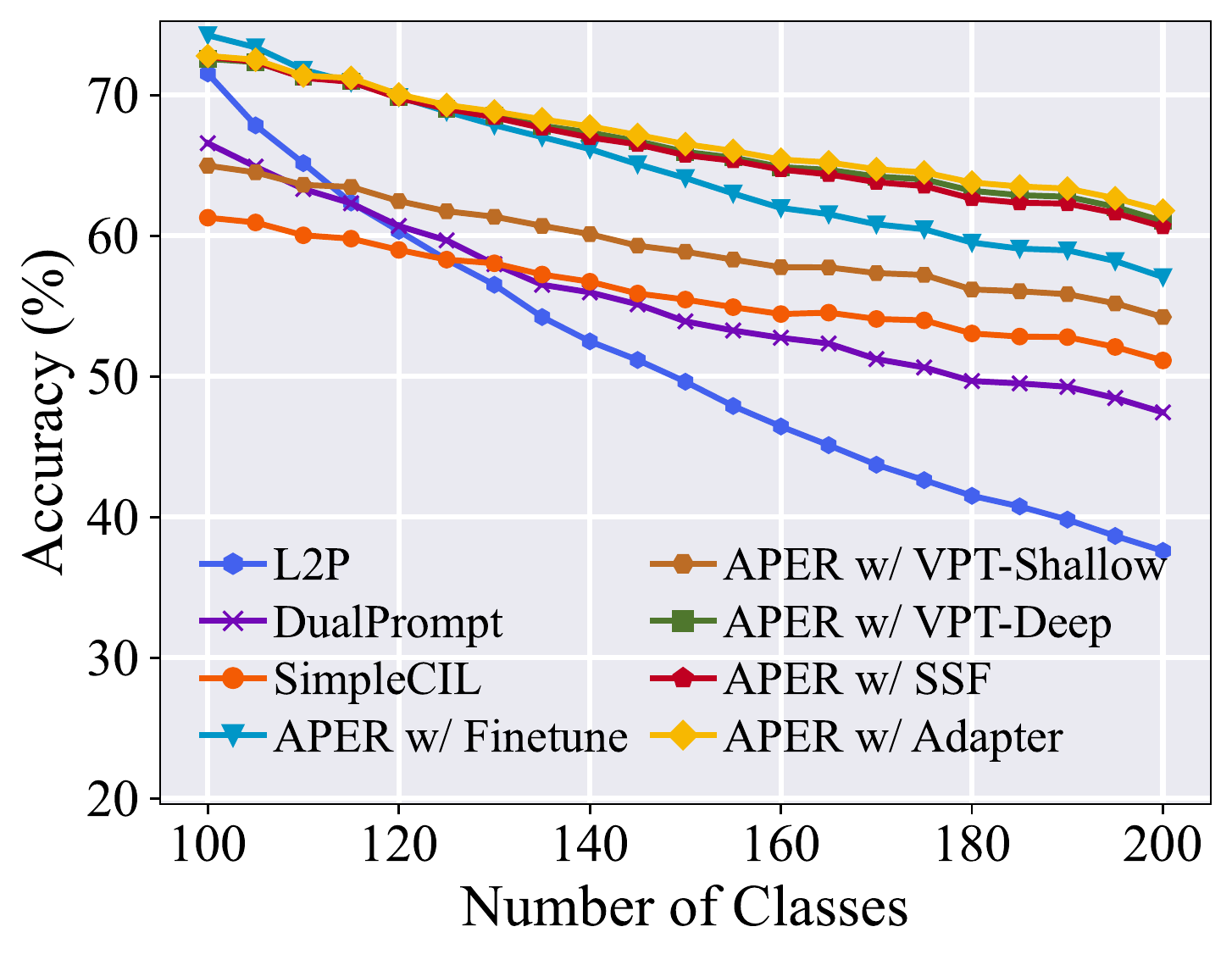}
			\label{figure:vit-obj}
		}
		\subfigure[CUB B0 Inc5]
		{\includegraphics[width=.65\columnwidth]{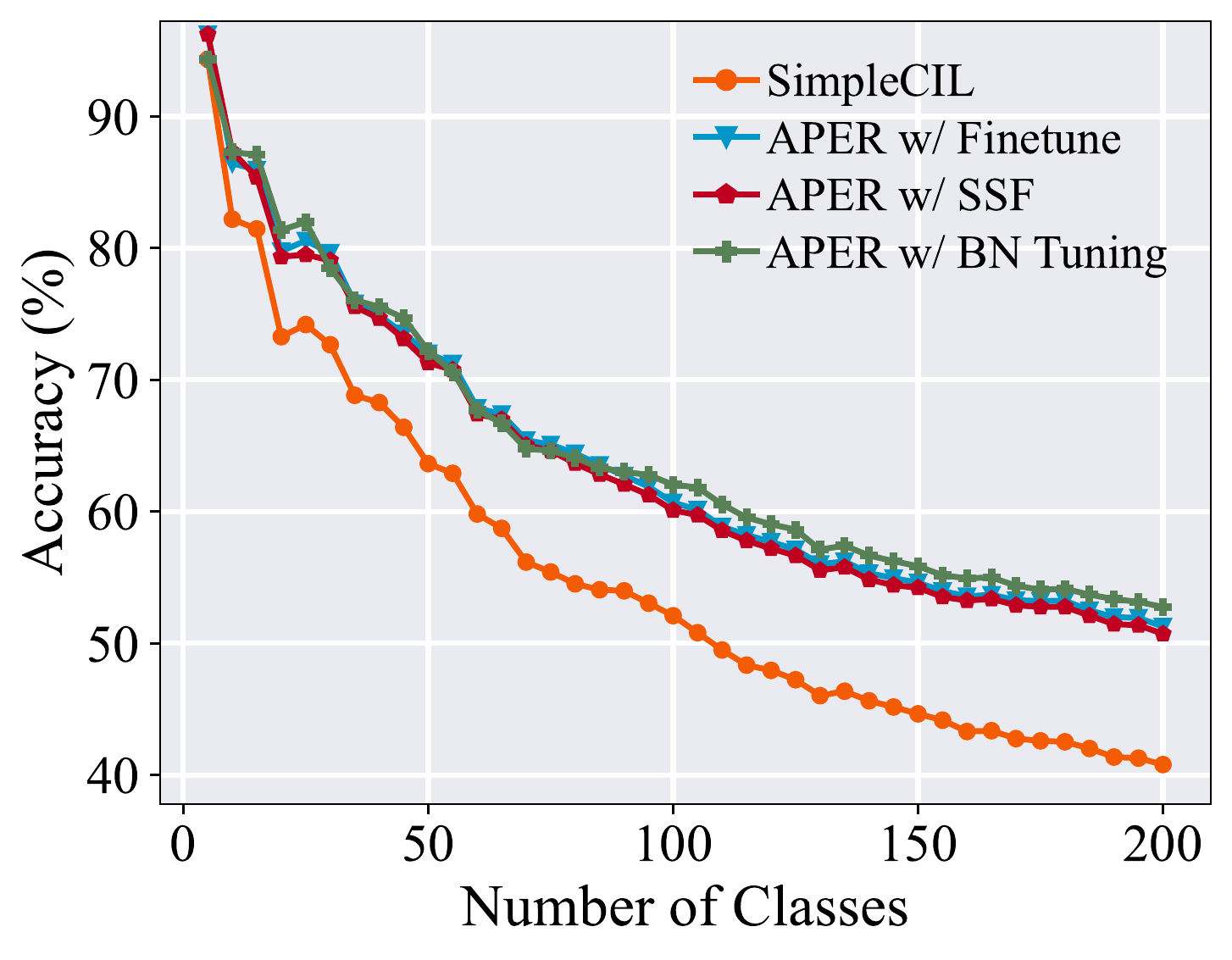}
			\label{figure:resnet-cub}
		}	
		\subfigure[ImageNet-A B0 Inc5]
		{\includegraphics[width=.65\columnwidth]{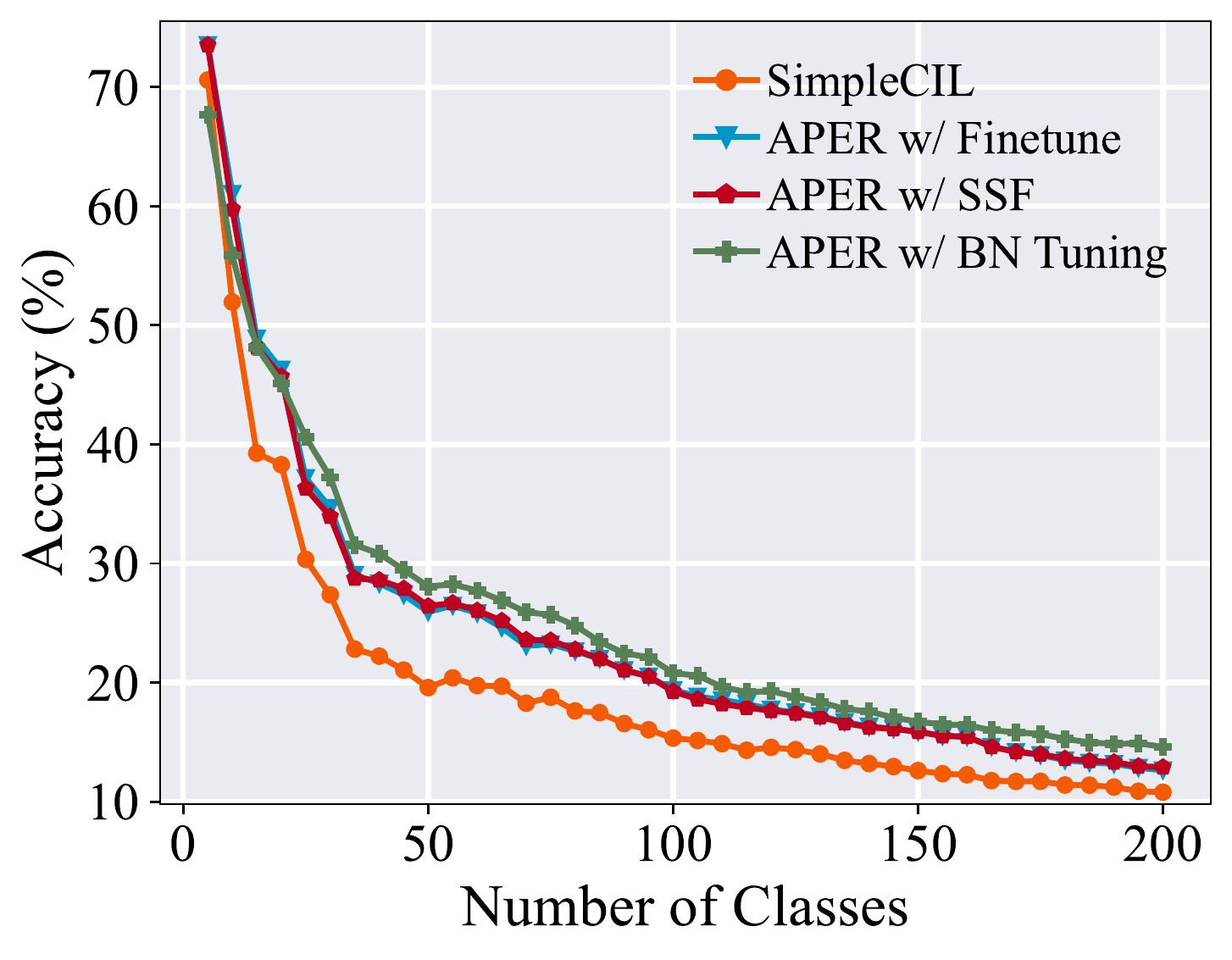}
			\label{figure:resnet-ina}
		}	
		\subfigure[CIFAR B0 Inc5]
		{\includegraphics[width=.65\columnwidth]{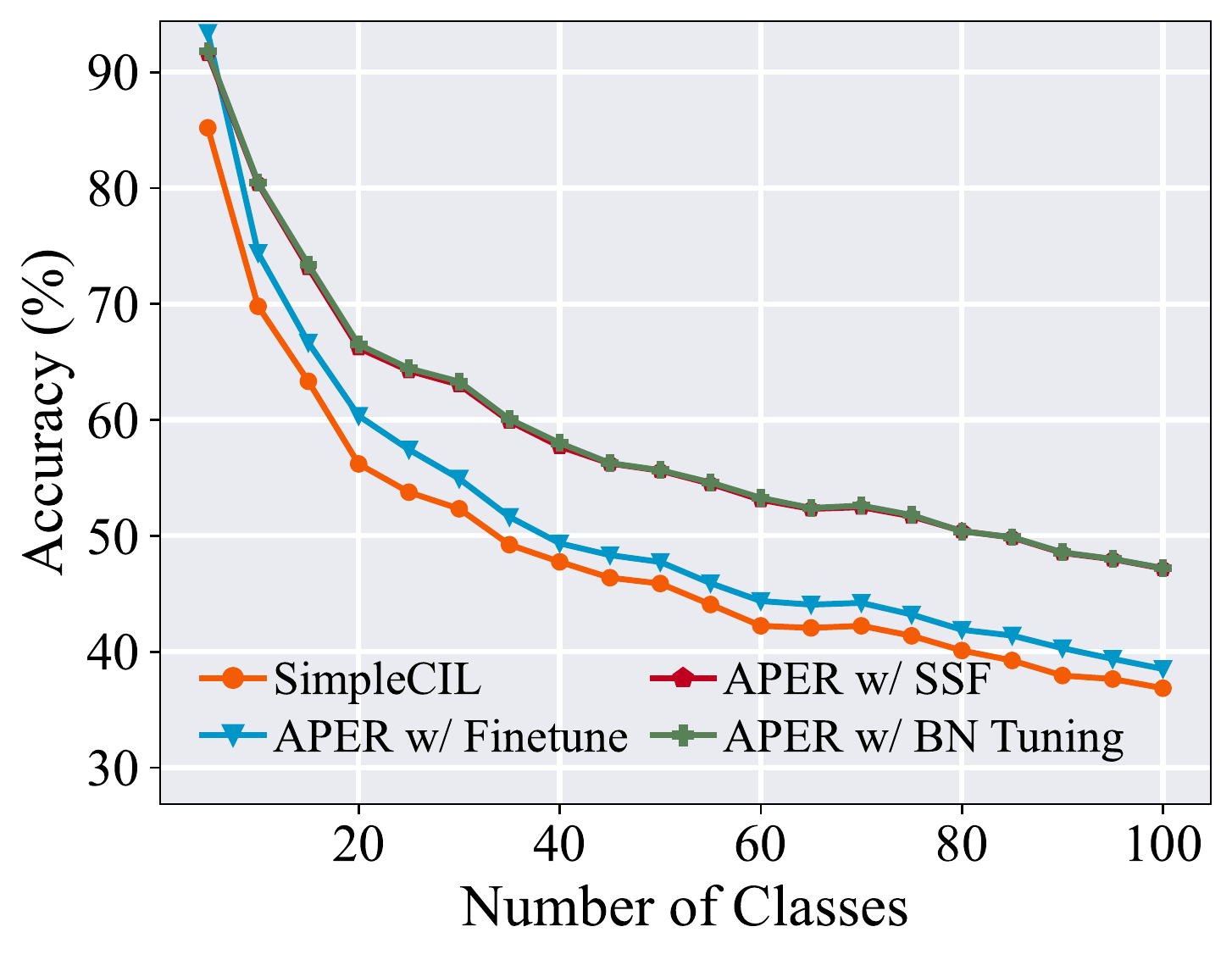}
			\label{figure:resnet-cifar}
		}	
		
	\end{center}
	\caption{ {\bf (a)$\sim$(f):} Incremental performance with {\bf ViT-B/16-IN1K} as the backbone when half of the total classes are base classes.
		{\bf (g)$\sim$(i):} Incremental performance when using {\bf ResNet18} as backbone. 
		Since L2P and Dualprompt cannot be deployed with ResNet, we do not report their performance in (g)$\sim$(i). \name consistently improves the performance of different backbones, \ie, ViT and CNN. 
	} 
	\label{figure:benchmark}
\end{figure*}	

\section{Experiments}

This section compares \name with state-of-the-art methods on benchmark datasets to show its superiority. Due to the overlap between pre-trained datasets and traditional class-incremental learning benchmarks, we also advocate four new benchmarks for evaluating PTM-based methods. Ablations and visualizations verify the effectiveness of \name with new classes. We also explore the performance of different pre-trained models in class-incremental learning.

\subsection{Implementation Details}
\noindent {\bf Dataset}: Following~\cite{wang2022dualprompt,yu2020semantic}, we evaluate the performance on CIFAR100~\cite{krizhevsky2009learning}, CUB200~\cite{WahCUB2002011}, and ImageNet-R~\cite{hendrycks2021many}. Since PTMs are often trained with ImageNet21K~\cite{deng2009imagenet}, evaluating PTM-based methods with ImageNet is meaningless. Hence, we advocate four new datasets that have {\em large domain gap} with ImageNet, namely ImageNet-A~\cite{hendrycks2021natural}, ObjectNet~\cite{barbu2019objectnet}, Omnibenchmark~\cite{zhang2022benchmarking} and VTAB~\cite{zhai2019large}. 
Among them, ImageNet-A and ObjectNet contain {\em challenging samples} that ImageNet pre-trained models cannot handle, while
Omnibenchmark and VTAB contain diverse classes from multiple {\em complex realms}.
To construct the CIL task, we sample 200 classes from ObjectNet and ImageNet-A, and 300 from Omnibenchmark. We sample 5 datasets from VTAB, each containing 10 classes, to construct the cross-domain CIL setting. The aim of using the subset is to ensure a simple split of training classes.
Following~\cite{rebuffi2017icarl}, we shuffle the classes with the same random seed and split them into
`B/Base-$m$, Inc-$n$.' It means the first dataset contains $m$ classes, and each following dataset contains $n$ classes. $m=0$ means the total classes are equally divided into each task.

\noindent {\bf Comparison methods:} We first compare to
SOTA PTM-based CIL methods L2P~\cite{wang2022learning}, DualPrompt~\cite{wang2022dualprompt}, CODA-Prompt~\cite{smith2023coda}, CPP~\cite{li2024steering}, and LAE~\cite{gao2023unified}. Additionally, we also modify classical CIL methods LwF~\cite{li2017learning}, SDC~\cite{yu2020semantic}, iCaRL~\cite{rebuffi2017icarl}, LUCIR~\cite{hou2019learning}, DER~\cite{yan2021dynamically}, FOSTER~\cite{wang2022foster}, MEMO~\cite{zhou2022model}, FACT~\cite{zhou2022forward} to {\bf utilize the  same PTM as the initialization}. Apart from SimpleCIL, we also report the baseline, sequentially tuning the model, denoted as Finetune.

\noindent {\bf Training details:}
We use PyTorch~\cite{paszke2019pytorch} and Pilot~\cite{sun2023pilot} to {\em deploy all models on Tesla V100 with the same network backbone}.
As there are various PTMs publicly available~\cite{rw2019timm}, we follow~\cite{wang2022learning,wang2022dualprompt} to choose the most representative ones, 
denoted as {\bf ViT-B/16-IN1K} and {\bf ViT-B/16-IN21K.} Both are pre-trained on ImageNet21K, while the former is additionally finetuned on ImageNet1K. 
During adaptation, we train the model with a batch size of $48$ for $20$ epochs and use SGD with momentum for optimization.
The learning rate starts from $0.01$ and decays with cosine annealing. The prompt length $p$ is $5$ for VPT, and the projection dim $r$ is $16$ for Adapter.

\noindent {\bf Evaluation protocol:}  Following~\cite{rebuffi2017icarl}, we denote accuracy after the $b$-th stage as $\mathcal{A}_b$. 
We use $\mathcal{A}_B$ (the performance after the last stage) and $\bar{\mathcal{A}}=\frac{1}{B}\sum_{b=1}^{B}\mathcal{A}_b$ (average performance along incremental stages) as measurements.

\subsection{Benchmark Comparison}

We report the performance against SOTA methods in Table~\ref{tab:benchmark}, where all methods are based on the pre-trained ViT-B/16-IN21K. We also train these models with pre-trained ViT-B/16-IN1K and show the incremental trend in Figure~\ref{figure:vit-cifar}$\sim$\ref{figure:vit-obj}. These data splits include settings with large and small base classes for a holistic evaluation.

Firstly, we can infer that the embeddings of PTMs are generalizable and can be directly applied for CIL to beat the SOTA. Specifically, the baseline SimpleCIL outperforms DualPrompt by {\bf 20\%} on CUB and {\bf 8\%} on ImageNet-A in terms of $\mathcal{A}_B$. 
However, strong PTMs can be further improved if they are adapted by \mame, as downstream tasks have a large domain gap with the pre-trained dataset. 
Specifically, we find \name {\em consistently} outperforms SimpleCIL in seven benchmark datasets. 
In contrast, sequentially finetuning the model suffers severe forgetting, which verifies the effectiveness of the adapt and merge protocol.
Specifically, L2P and DualPrompt suffer from forgetting due to prompts being overwritten in the latter stages and the linear layer's imbalanced weight norms.
Since \name only requires tuning the PTM in the first stage, it requires less training time and extra parameters than L2P and DualPrompt, as shown in Figure~\ref{figure:intro}.
Among the variations of adapting techniques, we find {\em SSF and Adapter are more efficient than VPT}. 
We also compare to state-of-the-art traditional CIL methods and modify their backbones into pre-trained ViT for a fair comparison. However, we can infer from Table~\ref{tab:benchmark-typicalmethods} that these methods work poorly without exemplars. Hence, \name achieves the best performance compared to various CIL algorithms under the fair comparison.

Apart from ViTs, \name also works well with pre-trained CNNs. We adopt the ImageNet1K pre-trained ResNet18~\cite{he2016deep} for evaluation and plot the incremental performance in Figure~\ref{figure:resnet-cub},\ref{figure:resnet-ina},\ref{figure:resnet-cifar}. Results show that \name consistently boosts the performance of pre-trained ViTs and CNNs. Specifically, we find a simple BN tuning technique achieves better performance than fully or partially finetuning the ResNet.

Lastly, as shown in Table~\ref{tab:benchmark}, the performance on typical benchmarks is approaching saturation as they have a small domain gap with ImageNet. By contrast, due to the large domain gap between our newly established benchmarks and ImageNet, there is still space for improvement, indicating the effectiveness and necessity of these new benchmarks.

\begin{figure*}[t]
	\begin{center}
		\subfigure[ PCA projected features]
		{\includegraphics[width=.657\columnwidth]{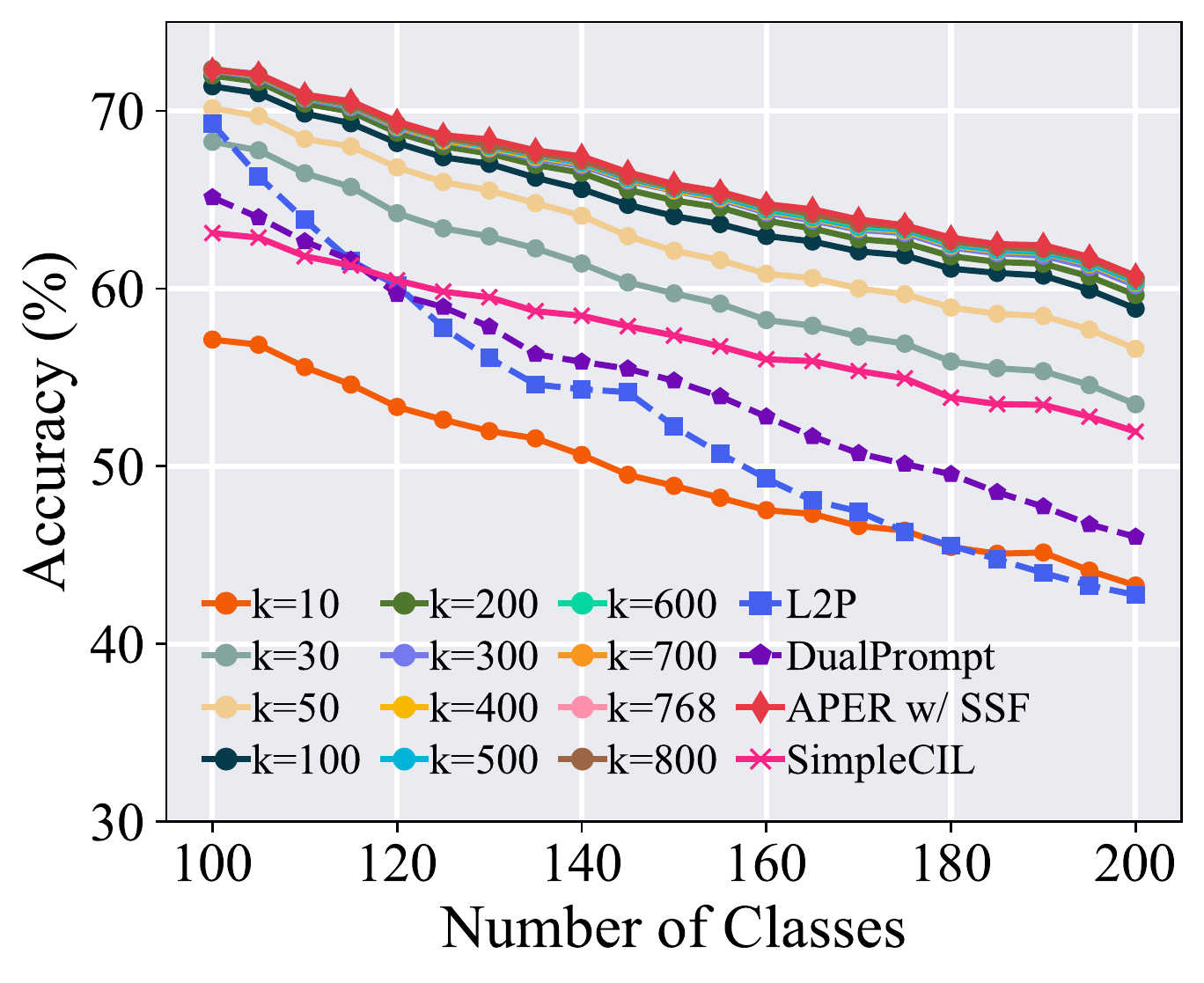}
			\label{figure:ablation-pca}}
		\hfill
		\subfigure[ Randomly sampled features]
		{\includegraphics[width=.657\columnwidth]{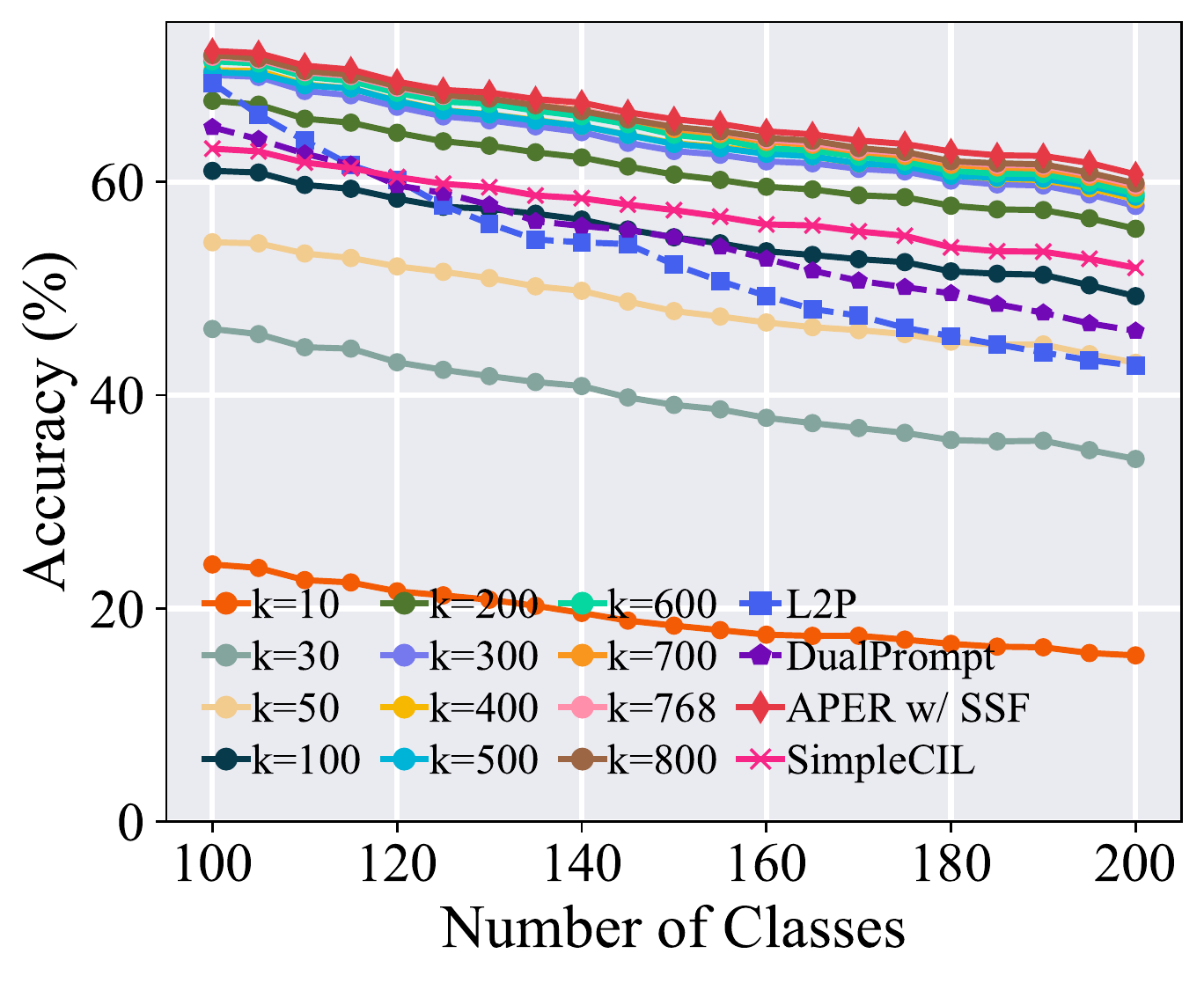}
			\label{figure:ablation-random}}
		\hfill
		\subfigure[ Projected dimension and accuracy]
		{\includegraphics[width=.657\columnwidth]{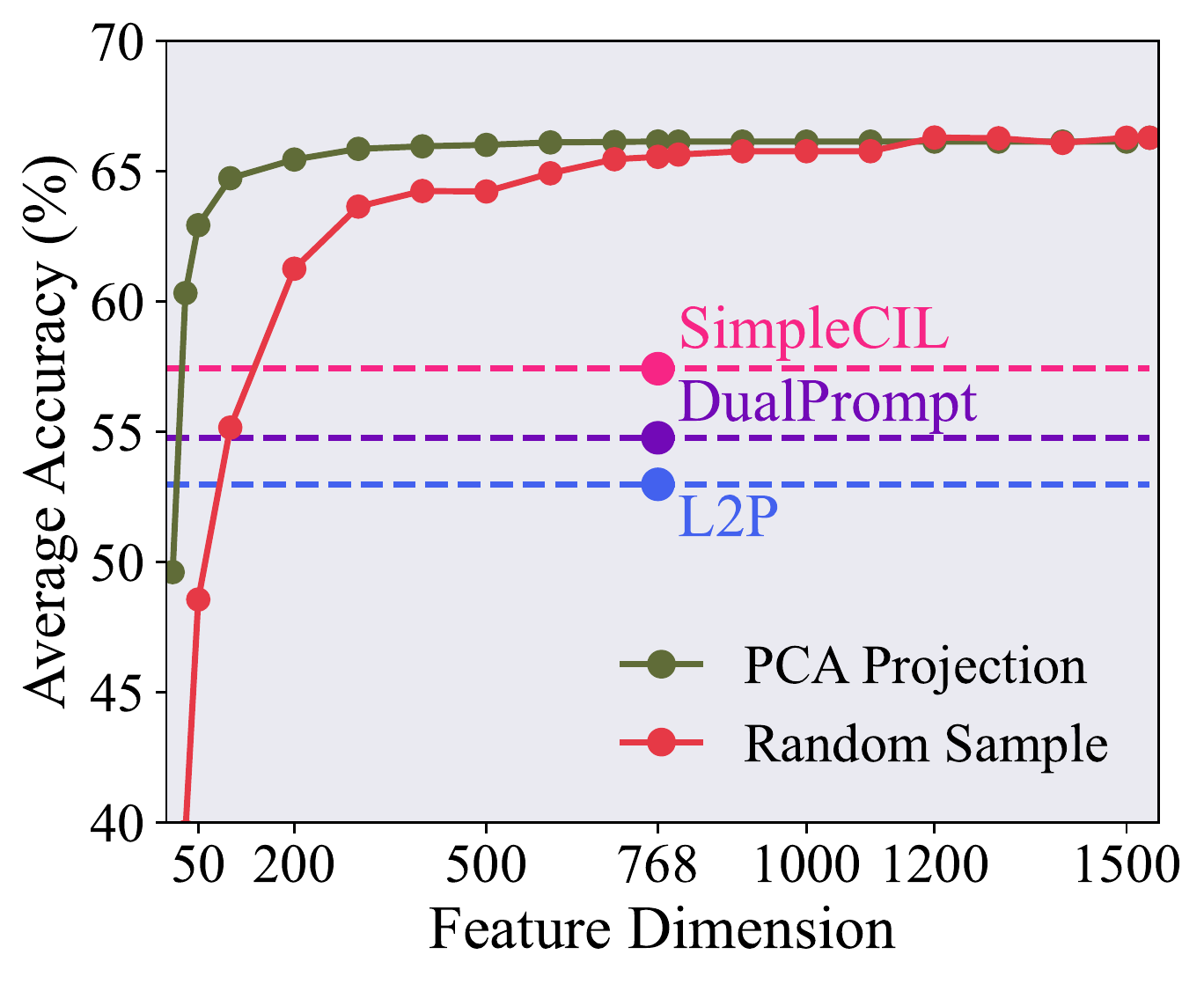}
			\label{figure:ablation-trend}	}\\
		\hfill	
		\subfigure[ Compositional components]
		{\includegraphics[width=.657\columnwidth]{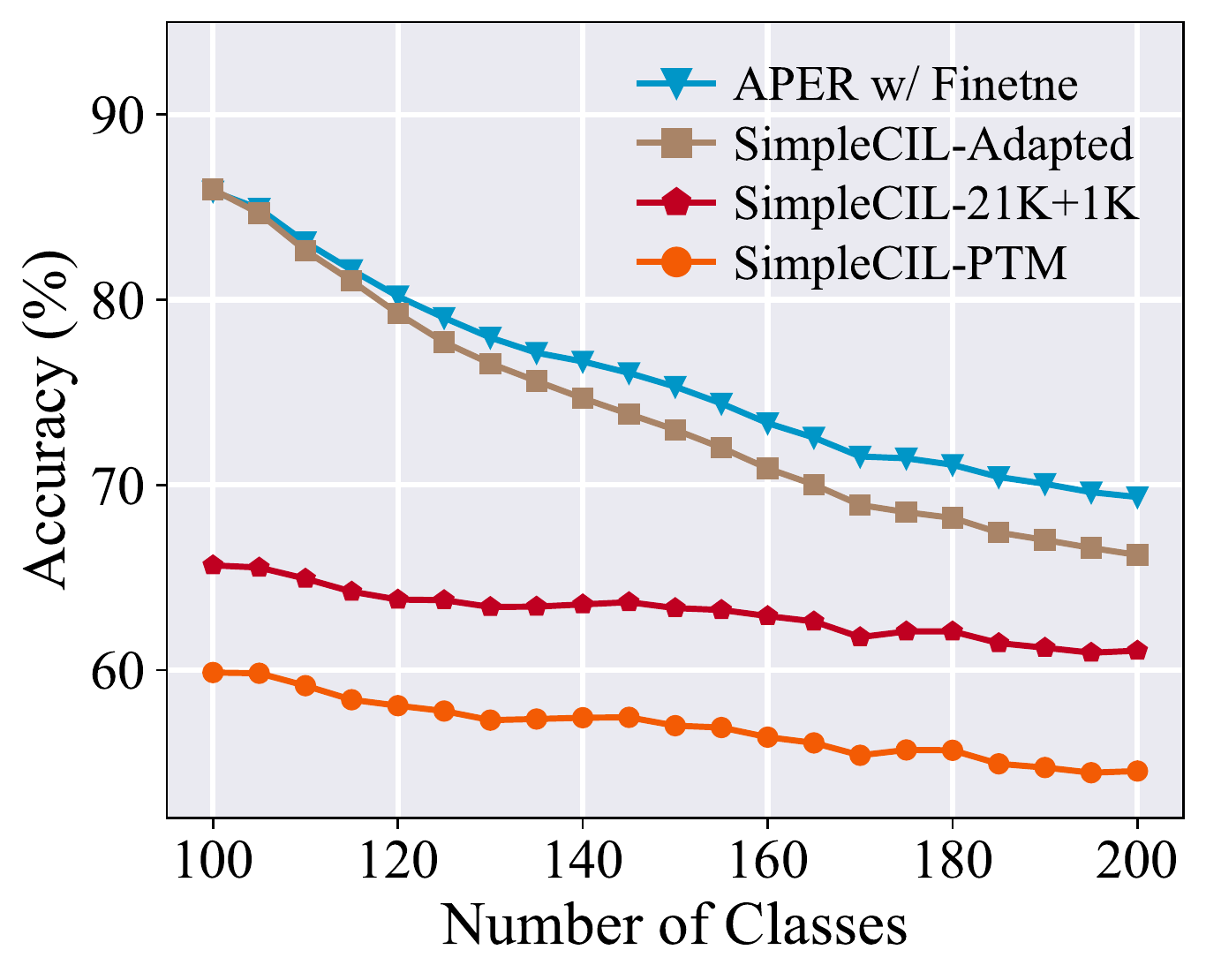}
			\label{figure:ablation-onevstwo}}
		\hfill
		\subfigure[ Number of parameters]
		{\includegraphics[width=.657\columnwidth]{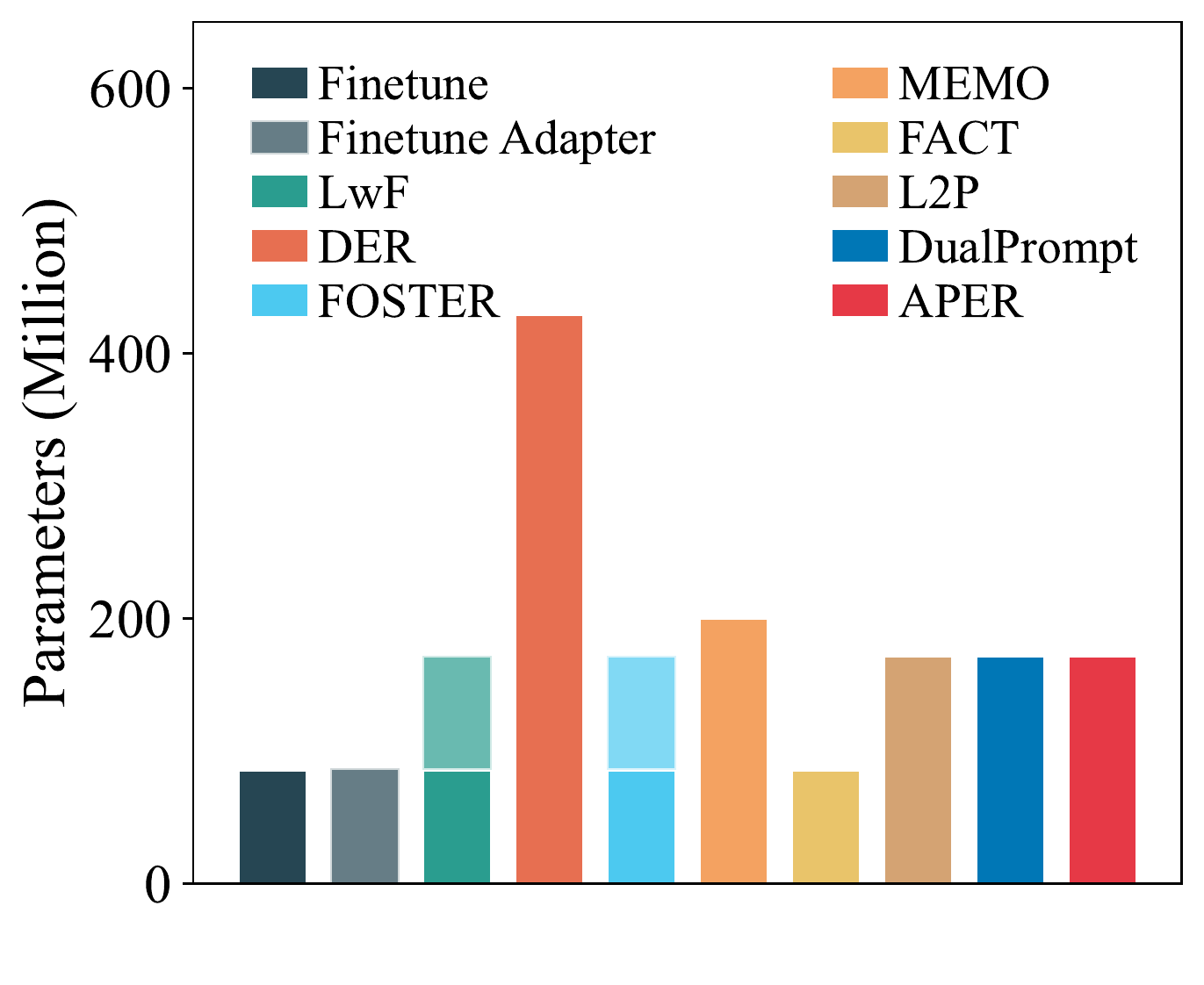}
			\label{figure:supp_param}}
		\hfill
		\subfigure[ Adapting stages]
		{\includegraphics[width=.657\columnwidth]{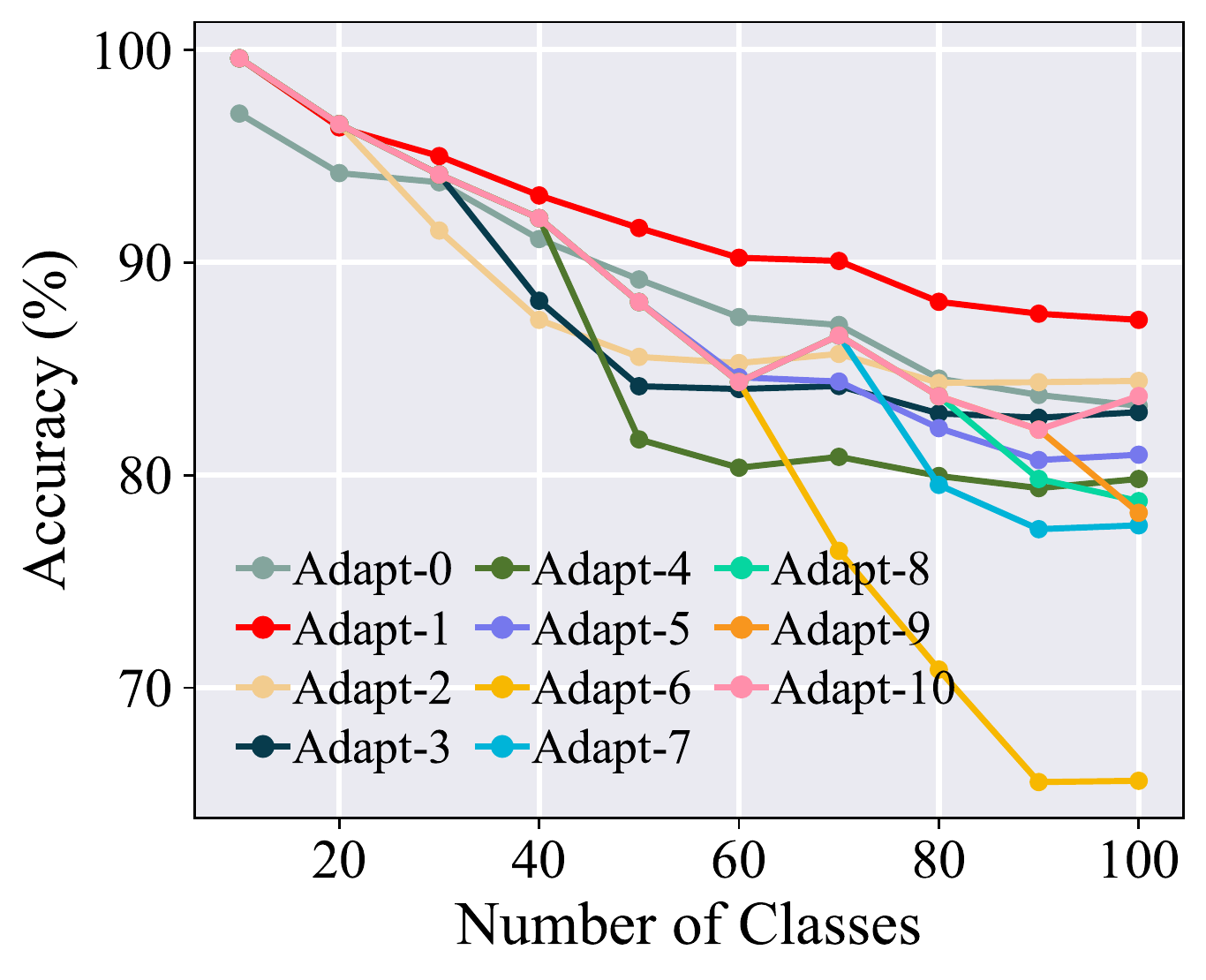}
			\label{figure:supp-generalizabilitya}}
	\end{center}
	\caption{\small Ablation study. {\bf (a)-(c)}: We use PCA or random sample to downscale the dimension of aggregated embeddings. {\bf (d)}: We compare \name to its sub-modules for ablation. {\bf (e)}: Number of total parameters of different compared methods. The bars with shadow denote the parameters used during training but dropped during inference. {\bf (f)}:  The accuracy trend with the change of adapting stages. Adapt-$T$ denotes the model is adapted for the first $T$ incremental tasks. $T=0$ denotes SimpleCIL. }
	\label{figure:analysis}
\end{figure*}

\begin{table}[t]
	\caption{\small Comparison to SOTA classical CIL methods with ViT-B/16-IN1K. {\em All methods are deployed without exemplars.} }
	\label{tab:benchmark-typicalmethods}
	\centering
		\begin{tabular}{@{}lcccccc}
			\toprule
			\multicolumn{1}{l}{\multirow{2}{*}{Method}} & 
			\multicolumn{2}{c}{ObjNet B0 Inc20} & \multicolumn{2}{c}{IN-A B0 Inc20}  \\
			& {$\bar{\mathcal{A}}$} & ${\mathcal{A}_B}$  
			& {$\bar{\mathcal{A}}$} & ${\mathcal{A}_B}$	\\
			\midrule
			iCaRL & 33.43 &19.18 &29.22 & 16.16\\
			LUCIR & 41.17 &25.89 & 31.09 &18.59\\
			DER   & 35.47  & 23.19  &33.85 & 22.27 \\
			FOSTER   & 37.83 & 25.07  &34.82 & 23.01\\
			MEMO & 38.52 &25.41 & 36.37 &24.46\\
			FACT & 60.59 & 50.96 & 60.13 & 49.82\\
			\midrule
			SimpleCIL &62.11 &51.13 & 59.67& 49.44\\
			\rowcolor{LightCyan}\name w/ SSF   & \bf68.75   &\bf  56.79 &\bf 63.59 &\bf  52.67 \\
			\bottomrule
		\end{tabular}
\end{table}

\subsection{Ablation Study}

In this section, we conduct an ablation study to investigate the influence of each part in \mame, \eg, using sampled features, part of its components, and different tuning stages. We also analyze the parameter number of different methods.

\subsubsection{Downscale features}
Since the feature of \name is aggregated with PTM and adapted model, it has a larger feature dimension than a vanilla PTM (\eg, 1536 versus 768). We conduct an ablation with \name w/ SSF on ObjectNet B100 Inc5 to show whether these features are essential for CIL. Specifically, we train a PCA~\cite{pearson1901liii} model in the first stage to reduce embedding dimension for the following stages. Denote the target dimension as $k$, we train the PCA model $\text {PCA}([\phi^*(\x),\phi(\x)]):\R^d\to\R^k$, and append it to the feature extractor. Hence, the features and prototypes are projected to $k$ dimensions. We plot the performance with the change of $k$ in Figure~\ref{figure:ablation-pca}. Specifically, we find \name obtains competitive performance to DualPrompt (with 768 dims) even if the features are projected to $30$ dims.

Apart from the PCA projection, we also experiment by randomly sampling $k$ features from the original feature space and report the results in Figure~\ref{figure:ablation-random}. The conclusions are consistent with the former ones, showing that randomly sampling $200$ dimensions in the concatenated space achieves the same performance scale as DualPrompt. 
We show the accuracy-dimension curves in Figure~\ref{figure:ablation-trend}.

\subsubsection{Sub-modules}

Since \name is concatenated with PTM and adapted model, we conduct ablations on ImageNet-A Base100 Inc5 with ViT-B/16-IN21K to compare \name w/ Finetune and its sub-modules.
Specifically, we build SimpleCIL with $\phi(\cdot)$ and $\phi^*(\cdot)$, respectively, denoted as {\bf SimpleCIL-PTM} and {\bf SimpleCIL-Adapted}. The former represents the capability of PTM, while the latter stands for the power of the adapted model. Both are compositional modules in \mame. Besides, we build SimpleCIL based on concatenated pre-trained ViT-B/16-IN21K and ViT-B/16-IN1K, denoted as {\bf SimpleCIL-21K+1K}. It utilizes the aggregated features of two embedding functions, which has the same dimension as \mame.

As shown in Figure~\ref{figure:ablation-onevstwo},
{\bf SimpleCIL-PTM} performs the worst among all variations, indicating that although pre-trained features are effective and generalizable, it still requires extracting features of downstream tasks for better representation. In comparison, 
{\bf SimpleCIL-Adapted} outperforms SimpleCIL-PTM, indicating the importance of model adaptation and adaptivity. 
However, adapting the model also overwrites the high-level features, which reduces the model's generalizability. The adapted model suffers more extensive performance degradation than vanilla SimpleCIL, indicating the effect of generalizability in resisting forgetting. Finally, {\bf \name w/ Finetune} outperforms any of these sub-modules with the help of unified adaptivity and generalizability.

\subsubsection{Parameter scale}

The parameter scale is another core factor influencing CIL algorithms in real-world applications, \eg, edge devices or mobile phones. In this section, we compare the parameter scale of different methods to investigate the possibility of real-world applications. In the comparison, all methods are based on the pre-trained ViT. For methods requiring backbone expansion (\eg, DER, MEMO, and FOSTER), we also use pre-trained ViT as the initialization of new backbones. 

We list the total number of parameters of all compared methods in Figure~\ref{figure:supp_param}, which indicates that \name obtains better performance than the compared methods with the same scale or fewer parameters.
Since L2P and DualPrompt have prompt pools, they rely on another pre-trained ViT as the `retriever' to search for the instance-specific prompt. Hence, \name shares the same scale of total parameters as these methods. Additionally, since \name utilizes parameter-efficient tuning techniques to obtain the adapted model, most of the parameters in the adapted model are the same as the pre-trained weight. Hence, the memory budget of \name can be further alleviated, which we will explore in future works.

\subsubsection{Influence of Adapting Stages}

In \mame, we only adapt the pre-trained model in the first incremental stage with $\D^1$. There are two reasons: 1) Sequentially tuning the model will suffer catastrophic forgetting. 2) Since we utilize a prototype-based classifier, tuning the model with multiple stages will result in {\em incompatible} features between former and new prototypes.

In this section, we conduct an ablation to determine the influence of adapting stages and report the results in Figure~\ref{figure:supp-generalizabilitya}. We conduct the experiment on CIFAR100 Base0 Inc10  setting with pre-trained ViT-B/16-IN21K. There are 10 incremental stages in total. We denote the tuning stages as $T$ and train \name w/ Adapter for ablation. Specifically, we change the tuning stages among $\{0,1,2,\cdots,10\}$ to determine the influence on the final performance. 
In the first $T$ stages, we adapt the PTM incrementally with adapter and replace the classifier with prototypes. Afterward, in the $T$-th stage, we freeze the encoding functions and only extract prototypes for the following stages. $T=0$ denotes vanilla SimpleCIL. To prevent forgetting, we freeze the classifier weight of former classes when learning new classes. 

As shown in the figure, tuning the model with the first stage achieves the best performance among all settings. Specifically, multi-stage tuning harms generalizability and results in the incompatible features of former and new classes.

\begin{figure}[t]
	\centering
	{
		\includegraphics[width=0.99\columnwidth]{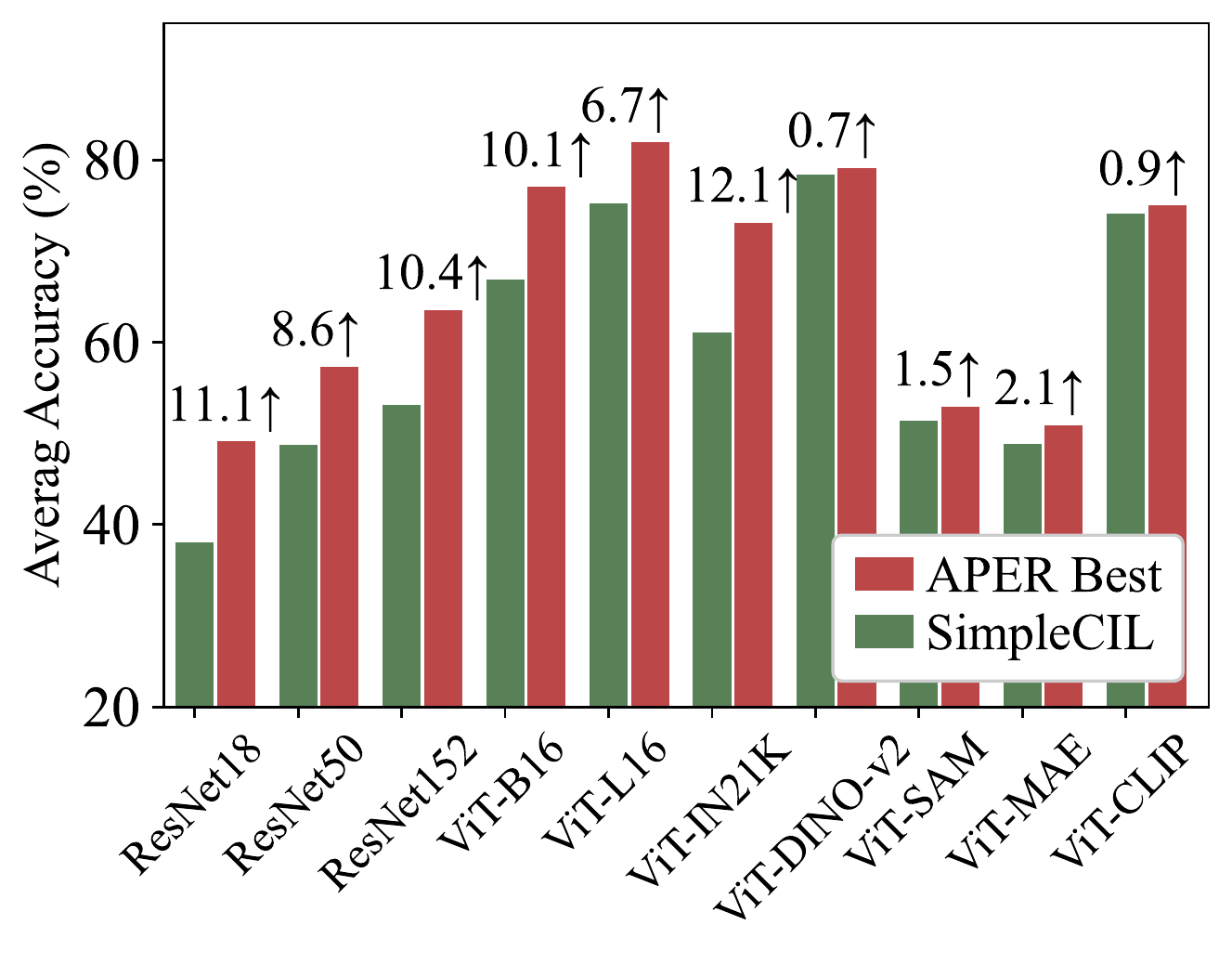}
	}
	\caption{\small  CIL with different kinds of PTMs on ImageNet-R Base0 Inc20. \name consistently improves the performance of different PTMs. We report the best variation of \name in the figure and its relative improvement above SimpleCIL in the figure.   }
	\label{figure:backbones}
\end{figure}

\subsubsection{Different PTMs}
Observing the performance gap between ViT-B/16-IN21K and ViT-B/16-IN1K, we seek to explore different kinds of PTMs on ImageNet-R Base0 Inc20. 
We choose publicly available PTMs, \ie, ResNet18/50/152~\cite{he2016deep}, ViT-B/16-IN1K/21K, ViT-L/16-IN1K, ViT-B/16-DINO-v2~\cite{oquab2023dinov2}, ViT-B/16-SAM~\cite{chen2022vision}, ViT-B/16-MAE~\cite{he2022masked}, ViT-B/16-CLIP~\cite{radford2021learning} (image encoder)  for a holistic evaluation, and report the results in Figure~\ref{figure:backbones}.
We can draw three main conclusions. 
Firstly, pre-trained ViTs show better generalizability than ResNets by achieving better performance with SimpleCIL. The main reason comes from the more extensive training data and parameters.
Secondly, larger ViTs generalize better than small ones, and ViTs trained with supervised loss perform better than unsupervised ones.
However, DINO-v2 shows the best performance due to its 1.2B training instances.
Thirdly, owing to the massive training corpus and the cross-modal information, CLIP performs better than ImageNet21K pre-trained ViTs. 
Finally, we find the best \name variation consistently improves the performance of SimpleCIL for any PTM, thus validating its effectiveness.

\begin{figure}[t]
	\begin{center}
		\subfigure[First stage]
		{\includegraphics[width=.467\columnwidth]{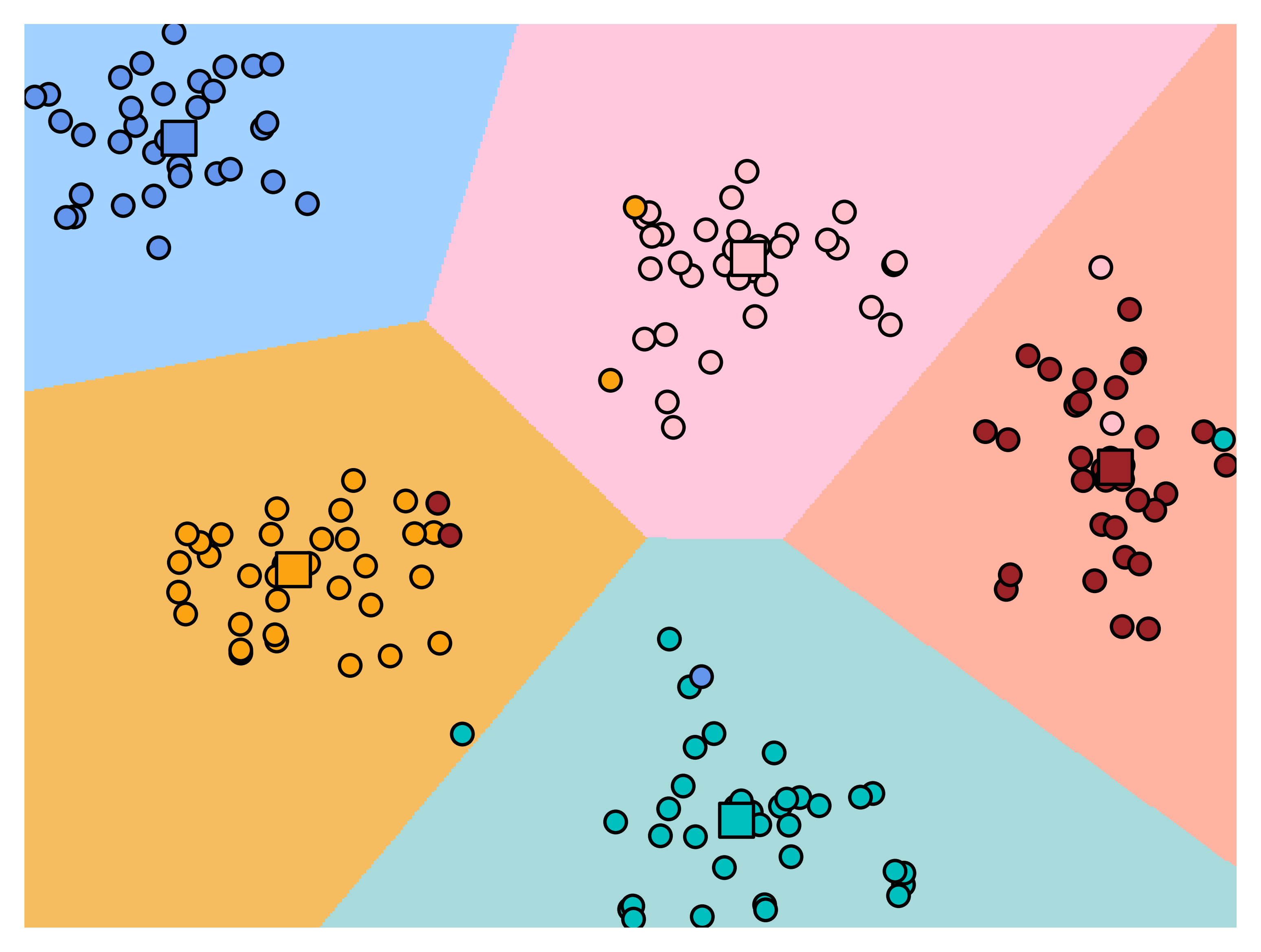}
			\label{figure:tsne1}}
		\subfigure[Second stage]
		{\includegraphics[width=.467\columnwidth]{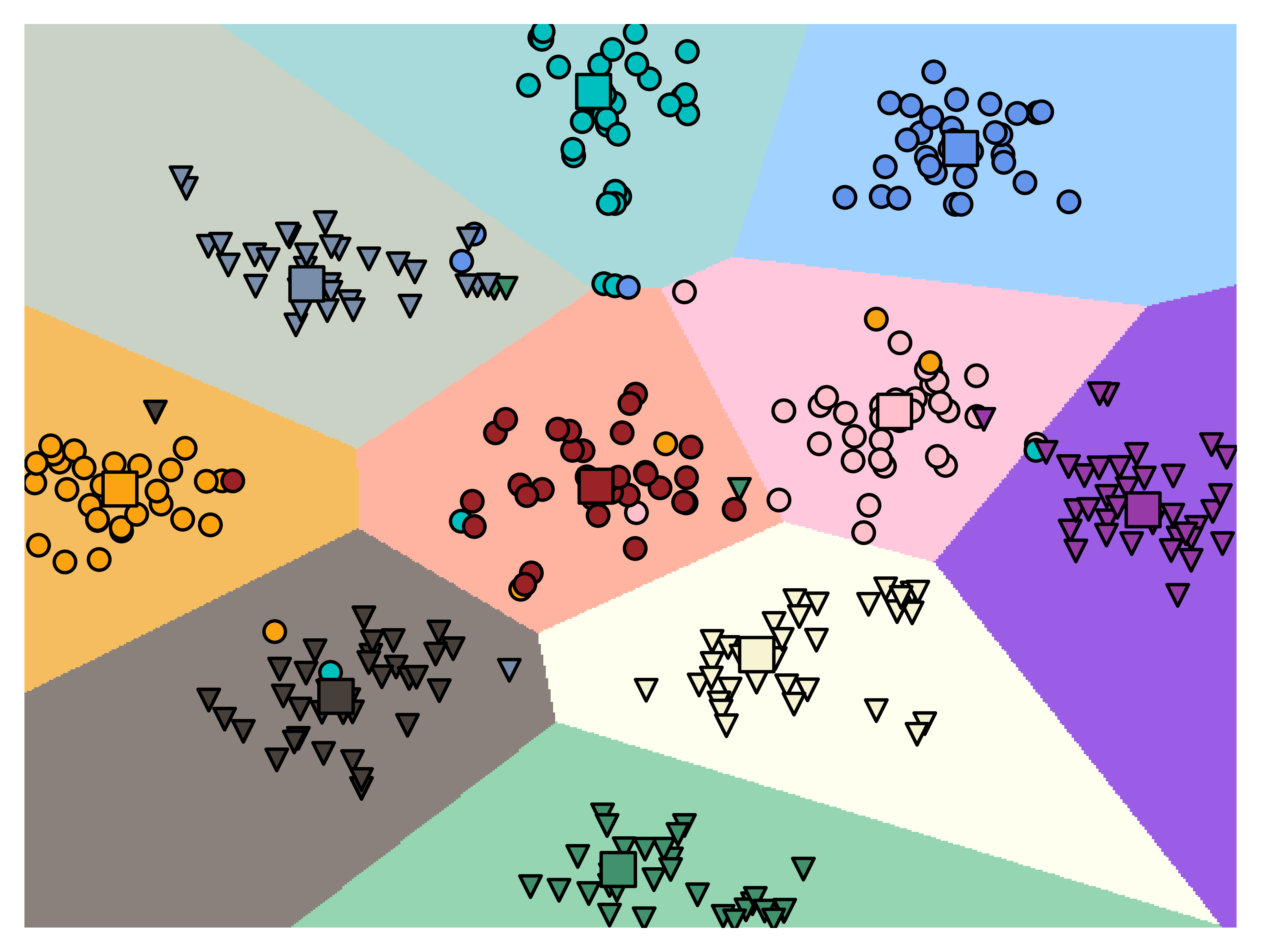}
			\label{figure:tsne2}
		}	
		\subfigure[Grad-CAM]
		{\includegraphics[width=.95\columnwidth]{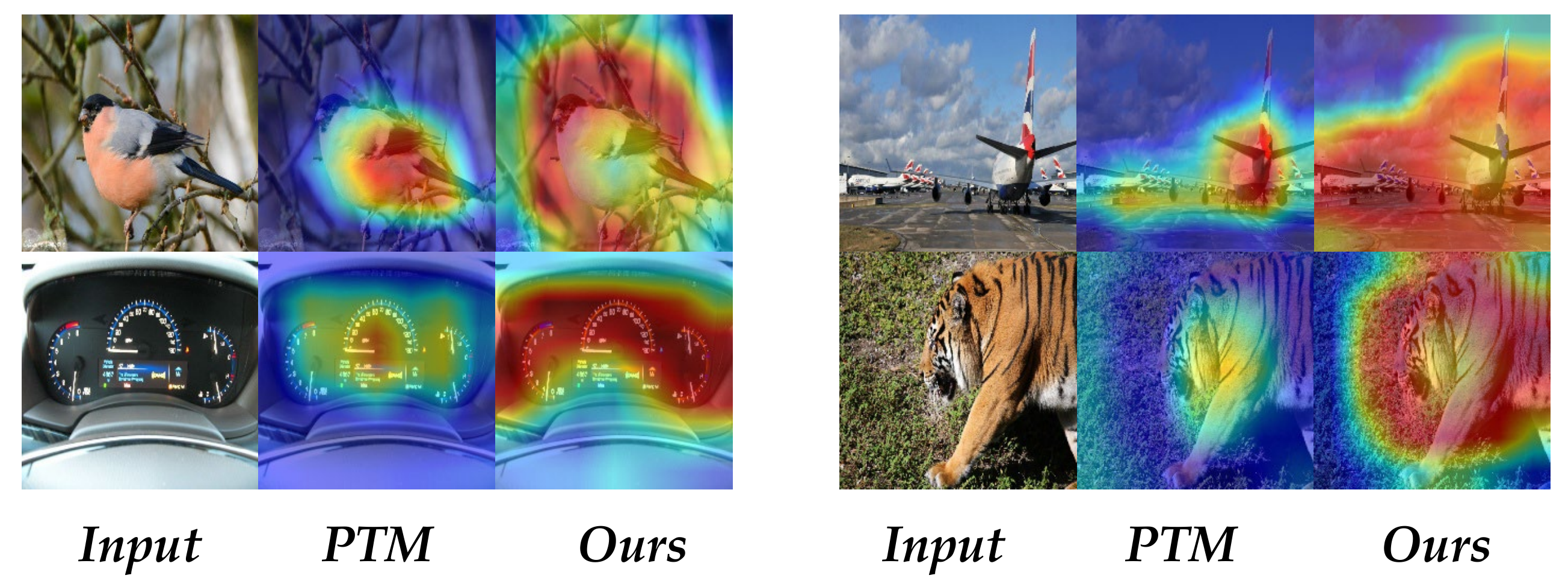}
			\label{figure:gradcam}
		}
	\end{center}
	\caption{ {\bf Top}: visualization of the  decision boundary on CIFAR100 between two incremental tasks. 
		Dots represent old classes, and triangles stand for new classes. Decision boundaries are shown with the shadow region. {\bf Bottom}: Grad-CAM visualizations of PTM and \mame. Important regions are highlighted with warm colors.
	}
	\label{figure:vis}
\end{figure}

\subsection{Visualization of Incremental Sessions}
In this section, we visualize the learned decision boundaries with t-SNE~\cite{van2008visualizing} on CIFAR100 dataset between two incremental stages, as shown in Figure~\ref{figure:tsne1}, \ref{figure:tsne2}. 
We visualize the classes from the first and second incremental tasks with colorful dots and triangles. Correspondingly, the class prototypes are represented by squares. As we can infer from these figures, PTM works competitively, which well separates the instances into their corresponding classes. The class prototypes are situated at the center of each class, verifying their representativeness in recognition. When extending the model from the first to the second stage, we find \name performs well on both old and new classes. Visualizations verify the generalizability and adaptivity of \mame.

We also visualize the Grad-CAM~\cite{selvaraju2017grad} results on OmniBenchmark dataset based on pre-trained ResNet18. Grad-CAM is utilized to highlight the critical regions in the image to predict the corresponding concept. The results are shown in Figure~\ref{figure:gradcam}, indicating \name concentrates more on the task-specific features than vanilla PTMs. Hence, visualizations verify the importance of adaptivity in PTM-based class-incremental learning.

\begin{table}[t]
	\caption{ Experiments on domain-incremental learning and cross-domain CIL. All methods are implemented with pre-trained ViT-B/16-IN1K. }
	\label{tab:crossdomain-data}
	\centering
	\footnotesize
		\begin{tabular}{@{}lcccccc}
			\toprule
			\multicolumn{1}{l}{\multirow{2}{*}{Method}} & 
			\multicolumn{2}{c}{Office-Home} & \multicolumn{2}{c}{DomainNet}  \\
			& {$\bar{\mathcal{A}}$} & ${\mathcal{A}_B}$  
			& {$\bar{\mathcal{A}}$} & ${\mathcal{A}_B}$	\\
			\midrule
			L2P & 71.61 & 59.48 & 57.90 & 57.59  \\
			DualPrompt & 70.85  & 58.26  & 62.41 & 54.81 \\
			\midrule
			SimpleCIL & 72.18 & 75.85 & 54.80 & 58.14 \\
			\rowcolor{LightCyan}\name w/ Finetune   & 75.44 &  77.47 & 64.17 & 60.78\\
			\rowcolor{LightCyan}\name w/ VPT-Shallow   &  71.44  & 63.26 & 58.75 & 60.84 \\
			\rowcolor{LightCyan}\name w/ VPT-Deep   & 71.06 & 61.27  &  60.38 & 58.31 \\
			\rowcolor{LightCyan}\name w/ SSF   & \bf 80.81  &  \textbf{83.14} & 65.79 & 65.02 \\
			\rowcolor{LightCyan}\name w/ Adapter   &   72.46 &  63.73 & \bf 67.50 & \textbf{66.90} \\
			\bottomrule
		\end{tabular}
\end{table}

\subsection{Experiments with Multiple Domains}

This paper mainly considers the CIL setting, where all the data are from the same domain. However, we also consider two challenging tasks to investigate the performance with significant domain gaps.
{\bf 1)}: Domain-incremental learning with Office-Home dataset~\cite{venkateswara2017deep}. Office-Home is a benchmark dataset for domain adaptation, containing four domains, each consisting of 65 categories. The four domains are: art, clip-art, product, and real-world, with an average of around 70 images per class and a maximum of 99 images in a class. We follow~\cite{wang2022learning} to organize the domain-incremental learning scenario where each task contains the classes of a new domain.
{\bf 2)}: Class-incremental learning with cross-domain data. We follow~\cite{smith2023coda} and split  
DomainNet~\cite{peng2019moment} into five tasks. Specifically, DomainNet	is a dataset of common objects in six different domains. All domains include 345 classes of objects. The domains include clip-art, real, sketch, infograph, painting, and quickdraw. To construct a class-incremental learning setting, we split the dataset into five tasks, each containing 69 classes of individual domains. 

The above settings contain datasets from multiple domains, and we report the experimental results in Table~\ref{tab:crossdomain-data}. We adopt the same pre-trained ViT-B/16-IN1K for all compared methods. Although \name is not specially designed with multiple stages, it still shows competitive performance against L2P and DualPrompt in these cross-domain tasks.

\section{Conclusion}

Learning with incremental classes is of great importance in real-world applications, which requires {\em adaptivity} for updating and {\em generalizability} for knowledge transfer. In this paper, we systematically revisit CIL with PTMs and draw three conclusions. Firstly, a frozen PTM can provide {\em generalizable} embeddings for CIL, enabling a prototype-based classifier to outperform the current state-of-the-art. Secondly, due to the distribution gap between pre-trained and downstream datasets, PTMs can be further harnessed to enhance their {\em adaptivity}. To this end, we propose \mame, which can be orthogonally combined with any parameter-efficient tuning method to unify generalizability and adaptivity for CIL.
Lastly, due to data overlapping, traditional ImageNet-based benchmarks are unsuitable for evaluation in the era of PTM. Hence, we propose four new benchmarks to evaluate PTM-based CIL methods. 
Extensive experiments verify \mame's state-of-the-art performance. 
Future work includes exploring task-specific tuning methods and structures. 

\noindent\textbf{Limitations}: The limitations are two-fold. 
Firstly, the model cannot make full use of exemplars since the adapted model should fully reflect the downstream features. It turns into exemplar-based CIL if sufficient old class instances are available, where adaptivity can be further addressed through data rehearsal. Secondly, since \name is only adapted with the first incremental stage, it shall face challenges when extensive domain gaps exist in the continual learning process, \ie, domain-incremental learning. In summary, extending \name to these more challenging scenarios are interesting future works.

\section{Acknowledgements}

This work is partially supported by National Science and Technology Major Project (2022ZD0114805), Fundamental Research Funds for the Central Universities (2024300373),
NSFC (62376118, 62006112, 62250069, 61921006), Collaborative Innovation Center of Novel Software
Technology and Industrialization, China Scholarship Council, Ministry of Education, Singapore, under its MOE AcRF Tier 2 (MOET2EP20221- 0012), NTU NAP, and under the RIE2020 Industry Alignment Fund – Industry Collaboration Projects (IAF-ICP) Funding Initiative.

\section{Data Availability Statement}

Code is available at \url{https://github.com/zhoudw-zdw/RevisitingCIL}.

\bibliographystyle{plain}      
\bibliography{paper}   

\end{document}